\documentclass[journal]{IEEEtran}

\usepackage{graphicx}
\usepackage{amsmath,amssymb} 
\usepackage{color}
\usepackage[noend]{algpseudocode}
\usepackage{multirow}
\usepackage{longtable}
\usepackage[T1]{fontenc}
\usepackage{epsfig}
\usepackage{authblk}
\usepackage[symbol*]{footmisc}
\usepackage{floatrow}
\usepackage{caption}
\usepackage{hyperref}
\usepackage{soul}
\soulregister\cite7 
\soulregister\citep7 
\soulregister\citet7 
\soulregister\ref7 
\soulregister\pageref7 

\newfloatcommand{capbtabbox}{table}[][\FBwidth]
\ifCLASSINFOpdf
\else
\fi
\begin{document}
%
\title{Attention Control with Metric Learning Alignment for Image Set-based Recognition}
%
%
%

\author{Xiaofeng~Liu,
        Zhenhua~Guo,
        Jane~You,
        and~B.V.K Vijaya~Kumar,~\IEEEmembership{Fellow,~IEEE}
\thanks{X. Liu, Zhenhua Guo, and B.V.K. Kumar are with the Department
of Electrical and Computer Engineering, Carnegie Mellon University, Pittsburgh,
PA, 15232 USA. Prof. Kumar is also with the Carnegie Mellon University Africa in Kigali, Rwanda.}
\thanks{J. You is with the Dept.of Computing, The Hong Kong Polytechnic University, Hong Kong.}
\thanks{Manuscript accepted Aug 05, 2019}}

%
%

\markboth{IEEE TRANSACTIONS ON INFORMATION FORENSICS AND SECURITY}%
{Shell \MakeLowercase{\textit{X. Liu et al.}}: Dependency-aware Attention Control for Image Set-based Face Recognition}
%



\maketitle

\begin{abstract}
This paper considers the problem of image set-based face verification and identification. Unlike traditional single sample (an image or a video) setting, this situation assumes the availability of a set of heterogeneous collection of orderless images and videos. The samples can be taken at different check points, different identity documents $etc$. The importance of each image is usually considered either equal or based on a quality assessment of that image independent of other images and/or videos in that image set. How to model the relationship of orderless images within a set remains a challenge. We address this problem by formulating it as a Markov Decision Process (MDP) in a latent space. Specifically, we first propose a dependency-aware attention control (DAC) network, which uses actor-critic reinforcement learning for attention decision of each image to exploit the correlations among the unordered images. {An off-policy experience replay is introduced to speed up the learning process}. Moreover, the DAC is combined with a temporal model for videos using divide and conquer strategies. {We also introduce a pose-guided representation (PGR) scheme that can further boost the performance at extreme poses. We propose a parameter-free PGR without the need for training as well as a novel metric learning-based PGR for pose alignment without the need for pose detection in testing stage.} Extensive evaluations on IJB-A/B/C, YTF, Celebrity-1000 datasets demonstrate that our method outperforms many state-of-art approaches on the set-based as well as video-based face recognition databases.
\end{abstract}


\IEEEpeerreviewmaketitle

\section{Introduction}

\IEEEPARstart{R}ecently, unconstrained face recognition (FR) has received much attention in computer vision research community \cite{liu2019research,liu2019feature,liu2018joint}. Initially, single image setting was standard for FR evaluations, $e.g.,$ Labeled Faces in the Wild (LFW) verification task \cite{huang2007labeled}. Increasing availability of cameras and digital storage has pushed the FR research into the next phase, where videos are available for face verification, $e.g.,$ YouTube Faces (YTF) dataset \cite{wolf2011face}. LFW and YTF datasets have a well-known frontal pose selection bias and the unconstrained FR is still considered an unsolved problem \cite{crosswhite2017template}.

In addition, open-set face identification is actually more challenging compared to the verification problem popularized by the LFW and YTF datasets \cite{liu2019permutation,liu2017line}. As shown in Fig. 1, comparison with $N$ gallery subjects is required which incurs $N$ times of computation of a single verification. 

\begin{figure}
\flushleft
~~~~~\includegraphics[height=3.55cm]{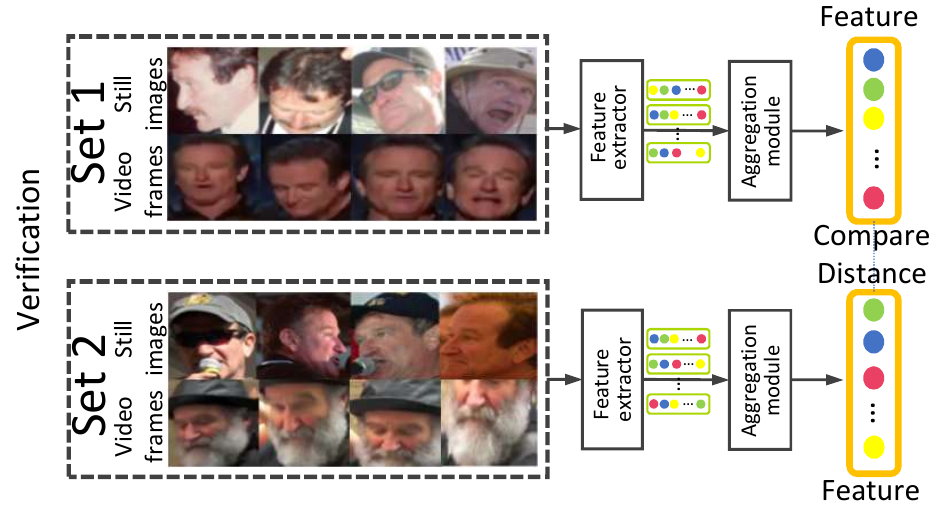}\\
~~~~~\includegraphics[height=3.5cm]{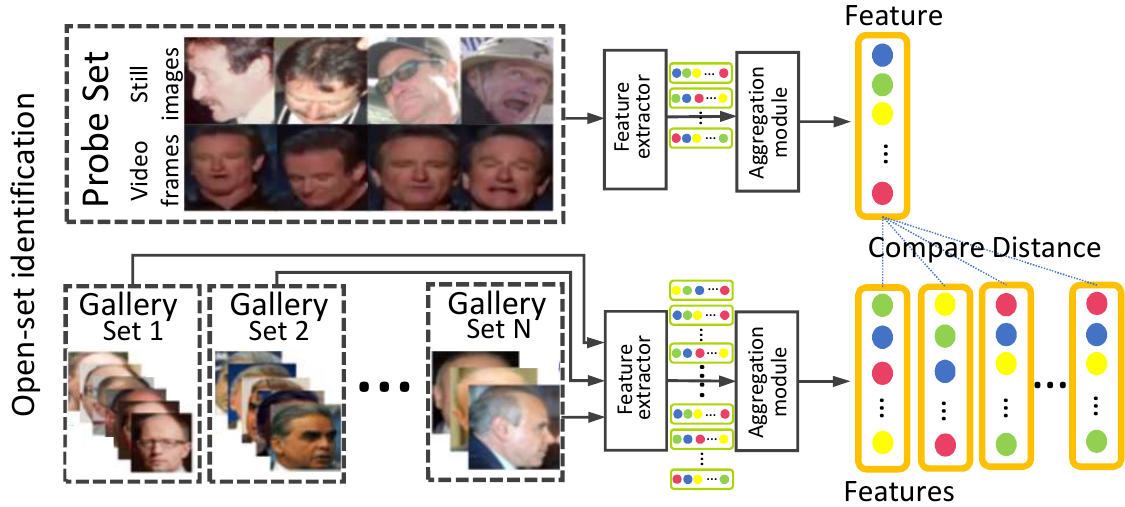}
\caption{ Illustration of the typical aggregation method for image set-based 1:1 face verification (above) and open-set 1:$N$ identification (bottom). Each set is independently represented as a single feature vector.}
\label{fig:ii1}
\end{figure}

The IARPA Janus Benchmarks (IJB-A \cite{klare2015pushing}, IJB-B \cite{whitelam2017iarpa}, IJB-C \cite{maze2018iarpa}) provide a more realistic unconstrained face verification and identification benchmark. They use a set (containing orderless images and/or videos with extreme head rotations, complex expressions and illuminations) as the smallest unit for face representation. The face set of a subject can be sampled from the mugshot history of that subject, the enrollment images for identity documents over lifetime, images taken at different check points, and the trajectory of a face in the video. This kind of setting is more similar to real-world biometric scenarios \cite{klare2015pushing}. {Capturing human faces from multiple views, background environments and cameras does result in large inner-set variations, but also provides more complementary information hopefully leading to higher accuracy in practical applications \cite{liu2017quality,liu2019conservative}.}

Considering the above factors, we propose to fully exploit both the inner- and inter-set relationship for the unified set-based face verification and identification. 

\begin{figure*}[t!]
\centering
\begin{tabular}{cc}
\includegraphics[height=6cm]{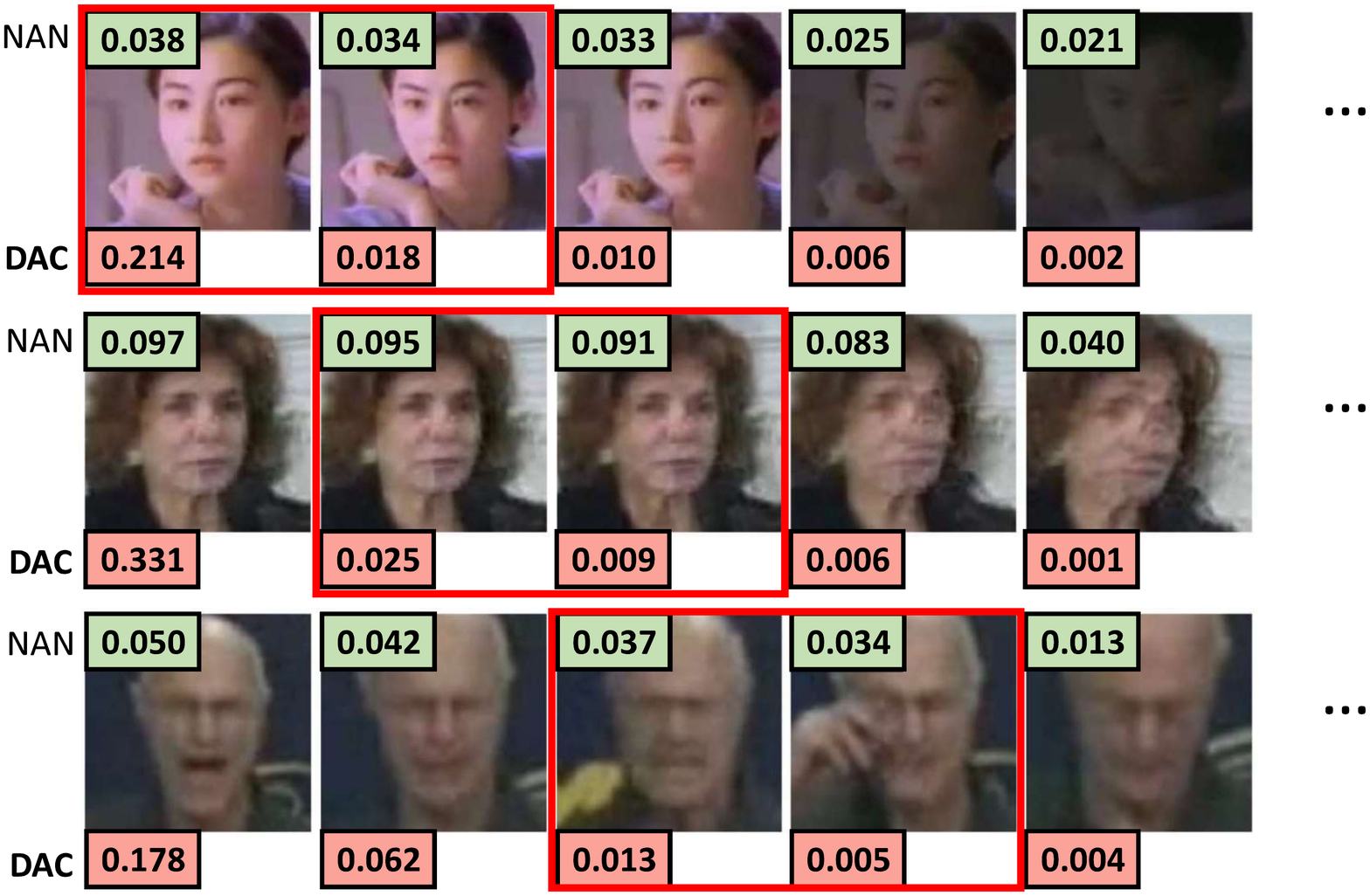}&~~~~~~~~~~~\includegraphics[height=6cm]{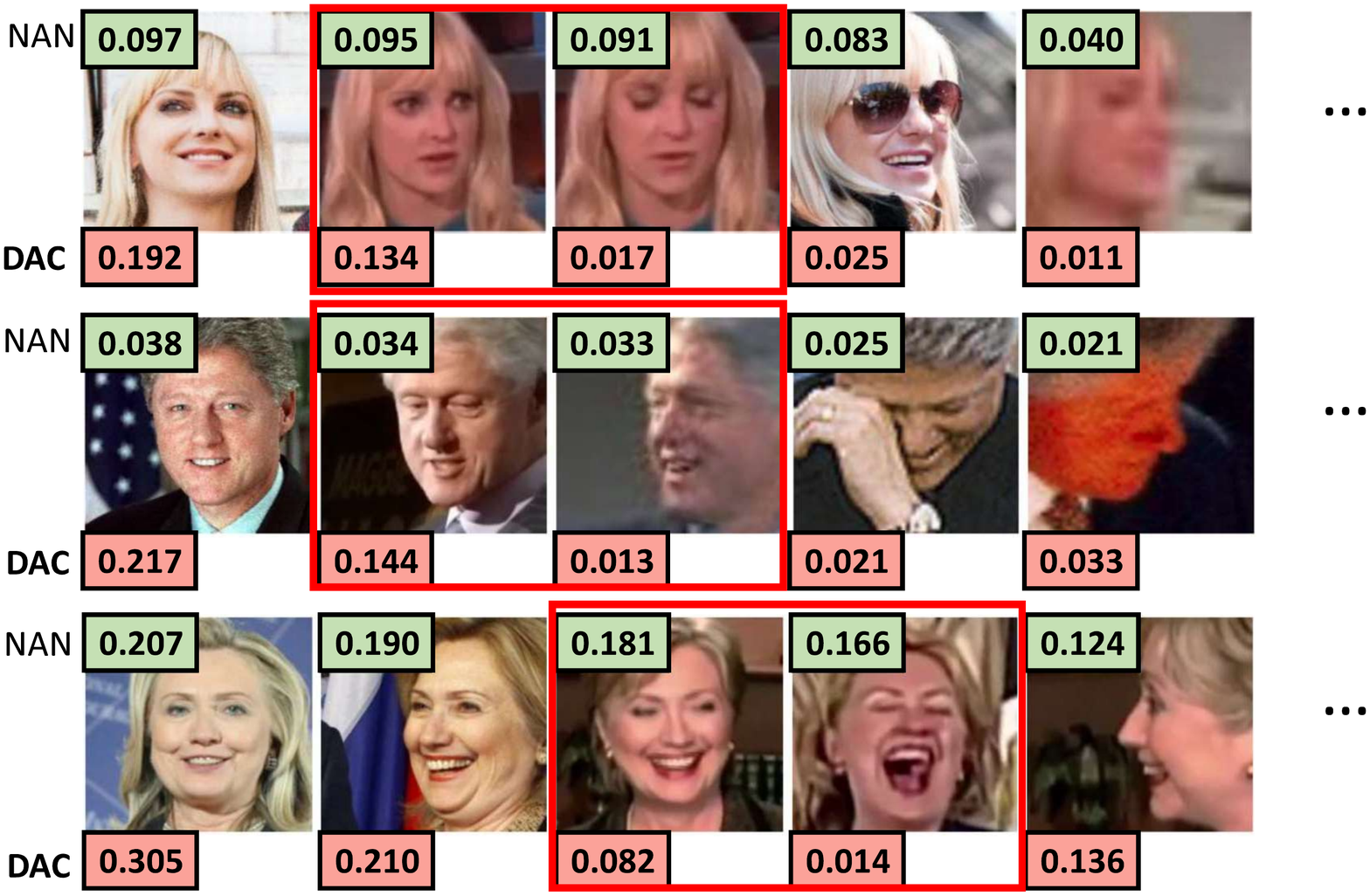}\\
(a)&~~~~~~~~~~~~~~~~~~~~(b)
\end{tabular}
\caption{Typical examples in the test set of (a) YTF and (b) IJB-A dataset showing the weights ($i.e.,$ importance) of images calculated by the previous method NAN \cite{yang2017neural}, and proposed DAC. NAN usually gives similar scores for the image pairs in red box, while the low-quality one does not introduce more complimentary information. The proposed DAC considers their redundancy and give a lower weight for the low quality version.}
\label{fig:ii2}
\end{figure*}

A commonly adopted strategy to aggregate identity information from multiple images is the average/max pooling \cite{parkhi2015deep,chen2015end}. Since images vary in quality, a neural network-based assessment module (NAN) has been developed to independently assign a weight for each frame \cite{liu2017quality,yang2017neural}. Larger weights correspond to more important/useful images. When using this, the frontal and clear faces are favored by NAN. However, this may result in redundancy and the model may not take advantage of the diversity in a set. As shown in Fig. 2, these low-quality frontal images are given relatively large weights in a set, sometimes as large as the weight given to the most discriminative one. There is little additional information that can be extracted from the blurry version of the same pose, while the valuable profile views $etc.,$ are almost ignored by the system. We argue that the weighting should depend on the other images within that set.

{Based on that insight, we formulate the attention scheme as a Markov Decision Process and use the actor-critic reinforcement learning (RL) to harness model learning. The dependency-aware attention control (DAC) module learns a policy to decide the importance ($i.e.,$ the weight) of each image step-by-step while observing other images in that set. In this way, we adaptively aggregate the feature vectors into a highly-compact representation inside the convex hull spanned by them.} Our approach not only explicitly learns to emphasize high-quality images while de-emphasizing low-quality ones, but also considers the inner-set dependency to reduce redundancy and preserves the benefit of information diversity.

Moreover, extracting set-level invariant features is always challenging because of varying poses, illumination conditions, resolutions $etc.$ Some approaches aggregate the image-level pair-wise similarity scores of all image pairs in two compared sets \cite{wen2016discriminative}. If $n$ is the average number of images in a set, then this approach exhibits $\mathcal{O}(n^2)$ computational complexity per match operation and $\mathcal{O}(n)$ space complexity per set, which are not desirable. More recently, \cite{zhang2017multi,rao2017attention} proposed a trade-off between speed and accuracy of processing paired video inputs using value-based Q-learning methods. These configurations focus on verification, and do not scale well for large scale identification tasks \cite{yang2017neural}. Conventionally, feature extraction of the probe and gallery samples are independent processes \cite{liu2017sphereface}. 

We notice that pose variation is the primary challenge in the IJB-A/B/C datasets and real-world applications, and we have prior knowledge that the structures of frontal and profile faces are significantly different. {Therefore, we simply utilize a parameter-free pose-guided representation (PGR) scheme to model the inter-set dependency. It well balances the computation cost and utilization of complementary information. The metric learning-based PGR further remove the pose detection in the testing stage by training the feature extractor to distribute images in feature space according to their identity and pose.}

Preliminary versions of the concepts in this paper were published in the 2018 European Conference on Computer Vision \cite{liu2018dependency}. In this paper, we extend those basic concepts in the following ways:

(1) {We design a novel metric learning-based PGR to align the pose, which eliminates the need for pose detection in the testing stage.}

(2) Using a divide and conquer strategy, we investigate the possible combination of DAC with temporal attention models ($e.g.,$ Bidirectional-recursive neural network, temporal convolution) to process orderless images and sequential video frames, not only boosting the performance, but also accelerating the training and the testing.

(3) We conduct all experiments using the new architecture, test on more challenging datasets, and provides more comprehensive ablation studies. 

In summary, this paper makes the following contributions.

(1) To the best of our knowledge, this is the first effort to use deep actor-critic reinforcement learning (RL) in visual recognition.

(2) The DAC can potentially be a general solution to incorporate rich correlation cues among orderless samples. Its coefficients can be trained in a normal recognition training task given only set-level identity annotation, without the need of extra supervision signals ($e.g.,$ quality label). We also show that it is compatible with a temporal attention model. 

(3) {To further improve sample efficiency, trust region-based experience replay is introduced to speed up the training and achieve stronger convergence. }

(4) {The PGR scheme well balances the computational cost and information utilization at the extremes of pose variation taking advantage of the domain knowledge about human face. Also, pose detection can be removed by the metric learning.}

(5) The module-based feature-level aggregation also inherits the advantage of conventional pooling strategies $e.g.,$ taking a varied number of inputs as well as offering time and memory efficiency.

We show that our method leads to the state-of-the-art accuracy results on the challenging IARPA Janus face recognition benchmarks and also generalizes well in several video-based face recognition tasks, \textit{e.g.}, YTF and Celebrity-1000 \cite{liu2014toward}.

The rest of this paper is organized as follows. Section 2 briefly reviews related literature. Section 3 introduces in detail the proposed on/off-policy actor-critic model for inner-set dependency control. Section 4 proposes the parameter-free and metric learning-based pose-guided representation for inter-set interaction. Section 5 reports numerical results and ablation study of our framework design. Finally, Section 6 provides our conclusions.
\section{Related Work}

\noindent{\bf{Image set/video-based face recognition}} has been studied in recent years \cite{yang2017neural}. Different from the single image setting \cite{liu2018data,liu2019mutual,liu2018ordinal}, The multi-image setting in the set-based dataset is similar to the multiple frames in the video-base recognition task. However, unlike video datasets, the temporal structure within a set is usually disordered, and the inner/inter-set variations are more challenging \cite{klare2015pushing}. There are two kinds of conventional solutions, namely, manifold-based methods and image-based methods. In the first category, each set/video is usually modeled by a manifold, and the similarity or distance is measured at the manifold-level \cite{liu2018dependency}. Manifold-based methods usually cannot handle the large appearance variations seen in the unconstrained FR task. In image-based methods, pairwise similarities between probe and gallery images are used for verification \cite{taigman2014deepface,schroff2015facenet,wen2016discriminative}. The quadratic number of image pair matches make them not scale well for identification tasks. Yang $et~al.$ \cite{yang2017neural} proposed an attention model to aggregate a set of features to a single representation with an independent quality assessment module for each feature. Then, there is only one feature vector for both the probe and gallery set, and one match is required to measure the similarity of two sets. Reference \cite{rao2018learning} up-samples the aggregated features to an image, then feeds it to an image-based FR network. However, weighting decision for an image does not take the other images into account as done in our proposed DAC method.

\noindent{\bf{Reinforcement learning (RL)}} trains an agent to interact (by trial and error) with a dynamic environment with the objective to maximize its accumulated reward. Recently, deep RL with convolutional neural networks (CNN) achieved human-level performance in Atari Games \cite{mnih2015human}. The CNN is an ideal approximate function to represent the infinite state space \cite{li2017deep}. There are two main streams to solve RL problems: methods based on value function and methods based on policy gradient. The first category, $e.g.,$ Q-learning, is the common solution for discrete action tasks \cite{mnih2015human}, which select the action leading to the maximum output. The second category can be efficient for continuous action space \cite{lillicrap2015continuous}, which uses the output as probability to choose each action, but we do not know the value of taking that action. There is also a hybrid actor-critic approach in which the parameterized policy is called an actor, and the learned value-function is called a critic \cite{mnih2016asynchronous}. As it is essentially a policy gradient method, it can also be used for continuous action space.

{Policy-based and actor-critic methods usually suffer from low sample-efficiency, high variance and often converge to local optima \cite{sutton2000policy}, since they typically learn via on-policy algorithms \cite{williams1992simple}, which makes policy prediction based only on the current policy.} Even the Asynchronous Advantage Actor-Critic approach \cite{mnih2016asynchronous} requires new samples to be collected for each gradient step. This quickly becomes expensive, as the number of gradient steps to learn an effective policy increases with task complexity. Off-policy learning instead aims to reuse past experiences, $i.e.,$ uses all the policies from its previous iteration stored by action-replay method to predict the current policy. This is not directly feasible with conventional policy gradient formulations, despite it being relatively straightforward for value-based methods \cite{li2017deep}. {Hence, in this paper, we focus on combining the stability of actor-critic methods with the efficiency of off-policy RL, which capitalizes on recent advances in deep RL \cite{mnih2016asynchronous}, especially off-policy algorithms \cite{schulman2015trust,wang2017sample}. }

In addition to its traditional applications in robotics and control, RL has recently been successfully applied to a few visual recognition tasks. Mnih \textit{et al.} \cite{mnih2014recurrent} introduce the recurrent attention model to focus on selected regions or locations from an image for digit detection and classification. This idea is extended to identity alignment by iteratively removing irrelevant pixels in each image \cite{lan2017identity}. The value-based Q-learning methods are used for object tracking \cite{huang2017learning} and the video verification in a computationally efficient way by dropping inefficient probe-gallery pairs \cite{rao2017attention} or stopping the comparison after processing sufficient number of pairs \cite{zhang2017multi,janisch2017classification} . However, this will inevitably result in losing the information in unused pairs and is only applicable for verification. There has been little progress made in policy gradient/actor-critic RL for visual recognition. 

\noindent{\bf{Deep metric learning}} {learns a new feature space by feature transformation \cite{liu2019hard,cheng2018learning,cheng2018deep}. It is popular for verification tasks by constructing a tuple of samples and requiring the samples with the same class label to be close while the distance between samples with different labels to be above a threshold \cite{liu2017adaptive}. \cite{cheng2017duplex} propose a stacked autoencoder by layer-wisely imposing a metric learning regularization term on the neurons in the hidden layers. In here, we do not set the distance constraints according to the subject identity, instead we choose the pose class to align the probe and gallery set.}

\begin{figure*}[t!]
\centering
\includegraphics[height=7cm]{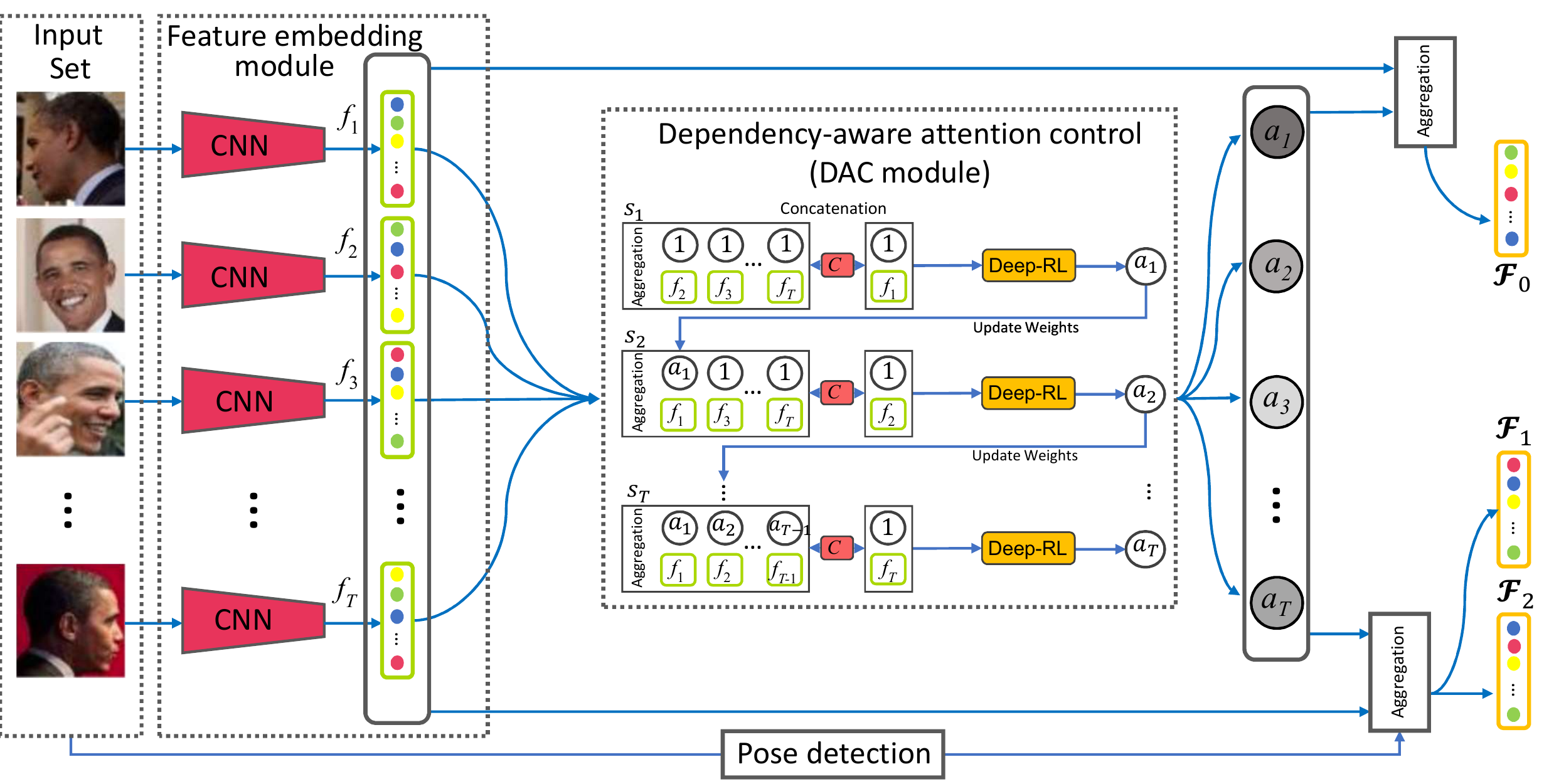}
\caption{ Our network architecture for image set-based face recognition.}
\label{fig:example}
\end{figure*}

\noindent{\bf{Temporal modeling}} is widely used in video classification, video-based identification and temporal action detection etc. \cite{zhou2017see} proposes to model the temporal information between frames using a recursive neural network (RNN). The 3D CNN is later developed to extract spatial-temporal features from video clips directly \cite{tran2015learning}. However, it is not compatible with our DAC module. Two types of temporal attention methods have been recently developed. The first method uses spatial convolution first followed by the fully connected (FC) layers \cite{liu2017quality}. The FC layers limit the model to fixed length video clip. The second approach chooses the temporal convolution instead of FC \cite{gao2018revisiting}. Although \cite{gao2018revisiting} only uses temporal convolution for action analysis with fixed length clip, we show that it is promising for videos-based face recognition with variable lengths.


\section{Inner-set dependency control}

The flow chart of our proposed framework is shown in Fig. 3. It takes a set of face images as input and processes them with two major modules to output a single(w/o PGR)/three(with PGR) feature vectors as its representation for recognition. We adopt a modern CNN module to embed an image into a latent space, which reduces the computation costs and offers a practicable state space for RL. Then, we cascade the DAC, which works as an attention model reading in all feature vectors and linearly combining them with adaptive weighting at the feature-level. Following the memory attention mechanism described in \cite{yang2017neural}, the features are treated as the memory and the feature weighting is cast as a memory addressing procedure. The PGR scheme can further utilize the prior knowledge of human face to address a set with large pose variation.

\subsection{Actor-critic approach for set-based face recognition}

In the set-based recognition task, we are given $M$ sets/videos $\mathop{({\mathcal{X}}^m, y^m)}_{m=1}^{M}$, where ${\mathcal{X}}^m$ is a image set/video sequence with varying number of images $T^m$ (\textit{i.e.,} ${\mathcal{X}}^m=\left\{{\mathop{x}_{1}^{m}},{\mathop{x}_{2}^{m}},\cdots,{\mathop{x}_{T^m}^{m}}\right\}$, ${\mathop{x}_{t}^{m}}$ is the $t$-th image/video frame in a set) and the $y^m$ is the corresponding set-level identity label. We feed each image ${\mathop{x}_{t}^{m}}$ to our model, and its corresponding feature representation ${\mathop{f}_{t}^{m}}$ is extracted using our neural embedding network. Here, we adopt the GoogLeNet \cite{szegedy2016rethinking} with Batch Normalization \cite{ioffe2015batch} or ResNet \cite{he2016deep} to produce a 128-dimensional feature vector as our encoding of each image. \cite{szegedy2016rethinking,he2016deep} have shown superior performance on several FR benchmarks, and can be easily replaced by other advanced CNNs. In the rest of the paper, we will simply refer to our neural embedding network as CNN, and omit the upper index (identity) where appropriate for better readability.

Since the features are deterministically computed from the images, they also inherit and display large variations. Hard attention scheme \cite{rao2017attention} that simply discards some of them may result in loss of too much information in a set \cite{yang2017neural}. Our attention control can be seen as the task of reinforcement learning to find the optimal weights of soft attention, which defines the importance of them in a weighted sum operation.

{Our solution to inner-set dependency modeling is to formulate it as a Markov Decision Process (MDP). At each time step $t$, the agent receives a state ${s_t}$ in a state space $\mathcal{S}$ and chooses an action ${a_t}$ from an action space $\mathcal{A}$, following a policy $\pi(a_t\mid s_t)$, which is the behavior of the agent. Then the action will determine the next state \textit{i.e.,} $s_{t+1}$ or termination, and receive a reward ${r_t(s_t,a_t)}\in \mathcal{R} \subseteq \mathbb{R}$ from the environment. The goal is to find an optimal policy $\pi^*$ that maximizes the discounted total return $R_t=\sum_{i\geq0}^{T}\gamma^{i}{r_{t+i}(s_t,a_t)}$ in expectation, where $\gamma\in [0,1)$ is the discount factor to trade-off the importance of immediate and future rewards \cite{li2017deep}.}

In the context of image-set based face recognition, we define the actions, $i.e., \left\{a^{1},a^{2},\cdots,a^{T}\right\}$, as the weights of each feature representation ${\mathop{\left\{f\right\}}_{i=1}^{T}}$. $T$ is the number of images (or frames) in a set. The weights of soft attention ${\mathop{\left\{a\right\}}_{i=1}^{T}}$ are initialized to be 1, and are updated step-by-step. The state $s_t$ is related to the $t-1$ weighted features and $T-(t-1)$ to-be-weighted features. In contrast to image-level dependency modeling, the compact embeddings largely shrink the state space and make our RL training feasible. In our practical applications, $s_t$ is the concatenation of $f_t$ and the aggregation of the remaining features with their updated weights at time step $t$. The termination means all of the images in this set have been successfully traversed.\begin{align}
{s_t}= \left\{\frac{(\sum_{i=1}^T {a_i}{f_i})-{f_t}}{(\sum_{i=1}^T {a_i})-1} \right\} Concatenate \left\{{f_t}\right\}
\end{align}

We define the global reward for RL by the overall recognition performance of the aggregated embeddings, which drives the RL network optimization. In practice, we add on top of the DAC few fully connected layers $h$ followed by a softmax to calculate the cross-entropy loss $L_m = -\log\left(\left\{e^{o_{y^m}}\right\}/{ \sum_{j=1}^M e^{o_j} }\right)$ to calculate the reward at this time step. We use the notation $o_j$ to mean the $j$-th element of the vector of class scores $o$. Let $g(\cdot)$ denote the weighted average aggregation function and $h$ maps the aggregated feature with the updated weights $g({\mathcal{X}}^m\mid s_{t})$ to $o$. The rewards are defined as follows:\begin{align}
g({\mathcal{X}^m}\mid s_{t},\text{CNN})=\sum_{i=1}^{T^m} \frac{{a_i}{f_i}^m}{\sum{a_i}}~~(\text{with~updated}~a_i~\text{at~step}~t)
\end{align}\begin{align}
{r_t}= &\left\{L_m[h(g({\mathcal{X}}^m\mid s_{t}))]-L_m[h(g({\mathcal{X}}^m\mid s_{t+1}))]\right\}\\\nonumber
&+\lambda\text{max}[0,({1-a_t})]
\end{align}where the hinge loss term serves as a regularization to encourage redundancy reduction and is weighted by $\lambda$. It also contributes to stabilizing training. The aggregation operation essentially selects a point inside of the convex hull spanned by all feature vectors \cite{cevikalp2010face}.

{Our original DAC traverses all of the images within a set to ensure that all of the samples are well considered. Since we initialize the weight of all images within a set to be 1, and assign a value $a\in[0,1]$ as the adjusted weight by DAC. These weights essentially eliminate redundant images. If we do not traverse all of the images, the un-traversed images will still have weight 1, which makes them appear very important, which is usually not true in real-world applications.}

{We can also propose another termination condition based on the results, i.e., termination occurs when the maximum value of softmax prediction reaches a threshold. That implies that the network is sufficiently confident of its prediction.}

{If the softmax prediction doesn’t reach a value larger than the threshold, then the method continues to traverse the images until all of the images in the set have been processed. We note that the reward or other metrics related to softmax prediction can be used for the training stage of open/closed-set identification and testing stage of closed-set identification. This is because we use softmax-based cross-entropy for the training of DAC and the closed-set identification has the same subject classes in its training and testing stage, which is a softmax-based multiclass classification problem. However, it cannot be applied to the testing stage of open-set identification, since we do not use softmax or calculate the reward. Note that open-set identification has different subject classes in its training and testing stages, which usually measure the similarity of probe feature with all of the gallery features as shown in Fig. 1.}

Considering that the action space here is a continuous space $\mathcal{A}\in\mathbb{R}^+$, the value-based RL ($e.g.,$ Q-Learning) cannot tackle this task. We adapt the actor-critic network to directly grade each feature dependent on the observation of the other features. In a policy-based method, the training objective is to find a parametrized policy ${\pi}_\theta(a_t\mid s_t)$ that maximizes the expected reward $J(\theta)$ over all possible aggregation trajectories given a starting state. Following the Policy Gradient Theorem \cite{sutton2000policy}, the gradient of the parameters given the objective function has the form:\begin{align}
 {\nabla}_\theta J(\theta)=\mathbb{E}[{\nabla}_\theta \text{log} {\pi}_\theta(a_t\mid s_t)({Q}(s_t,a_t)-b(s_t))] 
\end{align}

\noindent where ${Q}(s_t,a_t)=\mathbb{E}[R_t\mid s_t,a_t]$ is the state-action value function, in which the initial action $a_t$ is provided to calculate the expected return when starting in the state $s_t$. A baseline function $b(s_t)$ is typically subtracted to reduce the variance while not changing the estimated gradient \cite{williams1992simple,andrew1999reinforcement}. A natural candidate for this baseline is the state only value function ${V}(s_t)=\mathbb{E}[R_t\mid s_t]$, which is similar to $Q(s_t,a_t)$, except the $a_t$ is not given here. The advantage function is defined as ${A}(s_t,a_t)={{Q}(s_t,a_t)}-{{V}(s_t)}$ \cite{li2017deep}. Eq.(4) then becomes:\begin{align}
 {\nabla}_\theta J(\theta)=\mathbb{E}[{\nabla}_\theta \text{log} {\pi}_\theta(a_t\mid s_t){A}(s_t,a_t)] 
\end{align} 

\begin{figure}[t!]
\centering
\includegraphics[height=3.5cm]{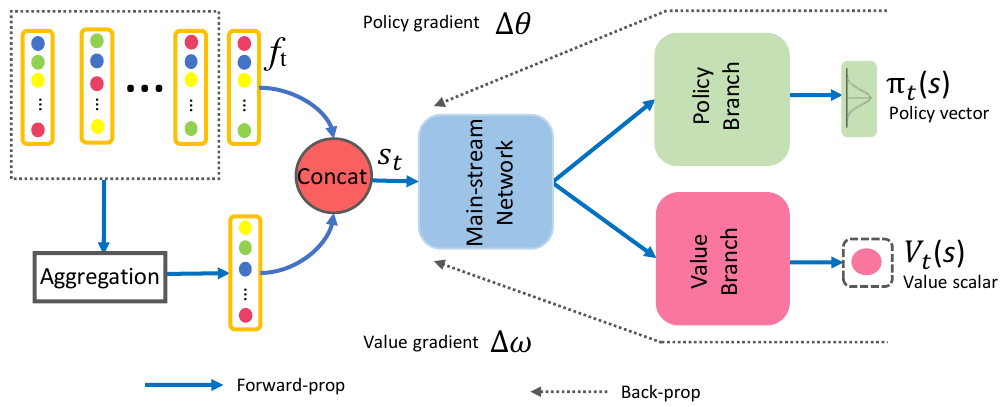}
\caption{The feed-forward neural network architecture of DAC(on/off) which uses the advantage actor-critic as the backbone.}
\label{fig:feed-forward}
\end{figure}

This can be viewed as a special case of the actor-critic model, where ${\pi}_\theta(a_t\mid s_t)$ is the actor and the ${A}(s_t,a_t)$ is the critic. To reduce the number of required parameters, the parameterized temporal difference (TD) error ${\delta_\omega}={r_t}+{\gamma}V_{\omega}(S_{s+1})-V_{\omega}(S_{s})$ can be used to approximate the advantage function. We use two different symbols $\theta$ and $\omega$ to denote the actor and critic function, but most of these parameters are shared in a main stream neural network, then separated to two branches for policy and value predictions, respectively. 

In the DAC we adopt two FC layers (100,100) then cascade two sub-branches with two-FC layers (64,16) as shown in Fig. 4. For all setups, we sample the learning rates log-uniformly in the range [$10^{-4}$, $10^{-3.3}$], and $\gamma=0.999$.

\subsection{Off-policy actor-critic with experience replay}

On-policy RL methods update the model with the samples collected via the current policy. Experience replay (ER) can be used to improve the sample-efficiency \cite{lin1992self}, where the experiences are randomly sampled from a replay pool $\mathcal{P} $. This ensures training stability by reducing data correlation. Since these past experiences were collected from different policies, the use of ER leads to off-policy updates. 

When training models with RL, $\varepsilon$-greedy action selection \cite{mnih2015human} is often used to trade-off between exploitation and exploration, whereby a random action is chosen with some probability and otherwise, the top-ranking action is selected. A policy used to generate a training weight is referred to as a behavior policy $\mu$, in contrast to the policy to-be optimized which is called the target policy $\pi$.

The basic advantage actor-critic (A2C) training algorithm described in Sec. 3.1 is on-policy, as it assumes that the actions are drawn from the same policy as the target to-be optimized (i.e., $\mu=\pi$). However, the current policy $\pi$ is updated with the samples generated from old behavior policies $\mu$ in off-policy learning. Therefore, an importance sampling (IS) ratio is used to re-scale each sampled reward to correct the sampling bias at time-step $t$: $\rho_t=\pi(a_t\mid s_t)/\mu(a_t\mid s_t)$ \cite{meuleau2000off}. For A2C, the off-policy gradient for the parametrized state only value function $V_\omega$ thus has the form:\begin{align}
\Delta\omega^{\text{off}}=\sum_{t=1}^{T}(\bar{R}_t-\hat{V}_\omega(s_t))\nabla_\omega\hat{V}_\omega(s_t)\prod_{i=1}^t {\rho_i} 
\end{align}

\noindent where $\bar{R}_t$ is the off-policy Monte-Carlo return \cite{precup2001off}:

\begin{align}
\bar{R}_t=r_t+\gamma r_{t+1}\prod_{i=1}^1 {\rho_i}+\cdots+\gamma^{T-t} r_{T}\prod_{i=1}^{T-t} {\rho_{t+i}}
\end{align}

Likewise, the updated gradient for policy $\pi_\theta$ is:\begin{align}
\Delta\theta^{\text{off}}=\sum_{t=1}^{T}\rho_{t} {\nabla}_\theta \text{log} {\pi}_\theta(a_t\mid s_t) \hat{\delta}_\omega
\end{align}where $\hat{\delta}_\omega=r_t+\gamma \hat{V}_\omega(s_{t+1}-\hat{V}_\omega(s_{t})${$)$} is the temporal difference (TD) error using the estimated value of $\hat{V}_\omega$.

Here, we introduce a modified Trust Region Policy Optimization method \cite{schulman2015trust,wang2017sample}. In addition to maximizing the cumulative reward $J(\theta)$, the optimization is also subject to a Kullback-Leibler (KL) divergence limit between the updated policy $\theta$ and an average policy $\theta_a$ to ensure safety. This average policy represents a running average of past policies and constrains the updated policy from deviating too far from the average $\theta_a\leftarrow[\alpha \theta_a+(1-\alpha)\theta]$ with a weight $\alpha$. Thus, given the off-policy policy gradient $\Delta\theta^{\text{off}}$ in Eq.(8), the modified policy gradient with trust region $z$ is calculated as follows:

\begin{equation}
     \begin{aligned}
{\mathop{}_{z}^{minimize}}&~~\frac{1}{2}\|\Delta\theta^{\text{off}}-z\|_2^2,\\
\text{Subject to:}& \nabla_\theta D_{KL}[\pi_{\theta_a}(s_t)\|\pi_{\theta}(s_t)]^{\text{T}} ~z \leq\xi  
\end{aligned}
\end{equation}

\noindent where $\pi$ is the policy parametrized by $\theta$ or $\theta_a$,
and $ \xi$ controls the magnitude of the KL constraint. Since the constraint is linear, a closed form solution to this quadratic programming problem can be derived using the KKT conditions. Setting $k=\nabla_\theta D_{KL}[\pi_{\theta_a}(s_t)\|\pi_{\theta}(s_t)]$, we get:

\begin{align}
z_{tr}^*=\Delta\theta^{\text{off}}-max\left\{  \frac{k^{\text{T}}\Delta\theta^{\text{off}}-\xi}{\|k\|_2^2},0\right\}k
\end{align}

This direction is also shown to be closely related to the natural gradient \cite{amari1998natural,peters2006policy}. {The above enhancements speed up and stabilize our A2C network training.} In our off-policy version of DAC, $\epsilon$ is sampled uniformly in the range [0.1,2], the truncation threshold $c$ is set to 5 and the rate of the soft updating $\alpha$ is set to 0.995.

\begin{figure}[t]
\centering
\includegraphics[height=3cm]{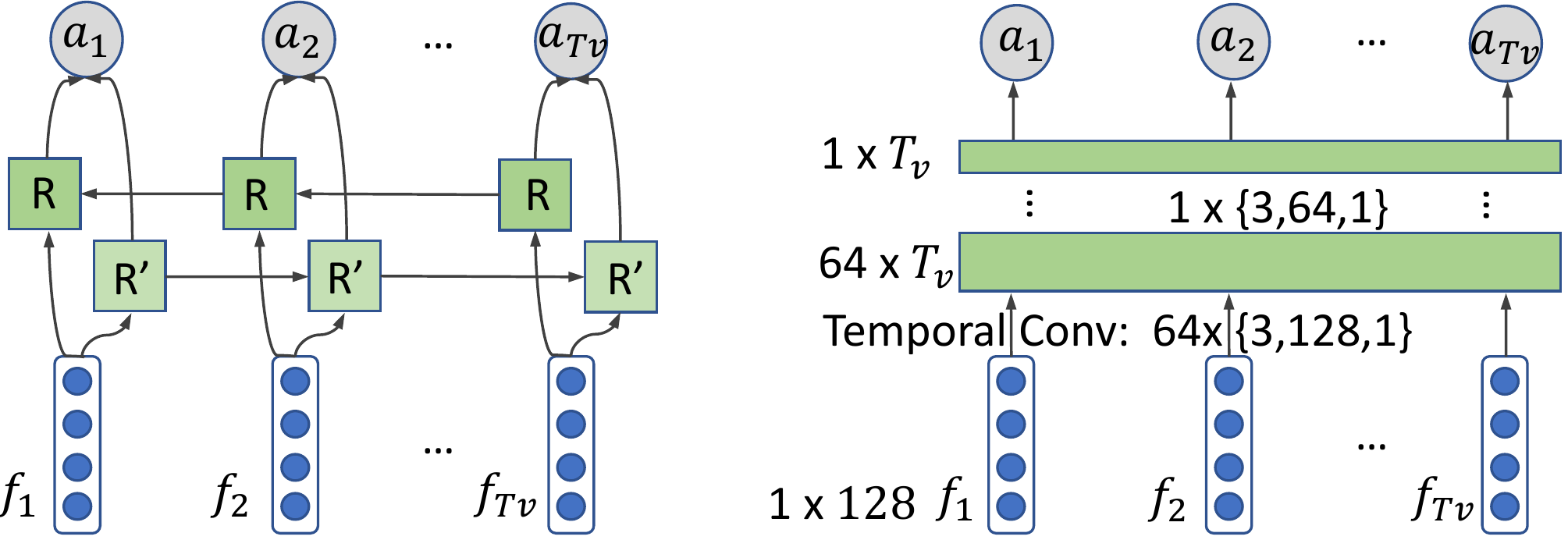}\\
(a)~~~~~~~~~~~~~~~~~~~~~~~~~~~~~~~~~~~~~~~~(b)
\caption{Illustration of the temporal attention scheme which takes $T_v$ continuous frames as input and outputs $T_v$ scalars as attention score. Left: recursive neural network-based temporal attention scheme. Right: spatial and temporal convolution-based attention scheme. R and R' are the neural networks, which look at some input features and output a scalar.}
\label{fig:mm5}
\end{figure}

\subsection{Combination with temporal attention}

Videos play an important role in a set. However, temporal dynamics presents in videos is ignored by both our preliminary version of DAC \cite{liu2018dependency} and conventional pooling methods \cite{yang2017neural}. Since video temporal analysis is a well-studied research area, we propose to process orderless images and sequential frames separately. 

Two possible video-based attention schemes are introduced to the set-based recognition. As shown in Fig. 5(a), the first scheme is based on an RNN. Specifically, the bi-directional long short-term memory (LSTM) is employed as the recurrent layer, which takes the sequential feature vectors as inputs and produces a sequence of activations. Then, the activations are normalized via the softmax.

\begin{figure}[t]
\centering
\includegraphics[height=2.7cm]{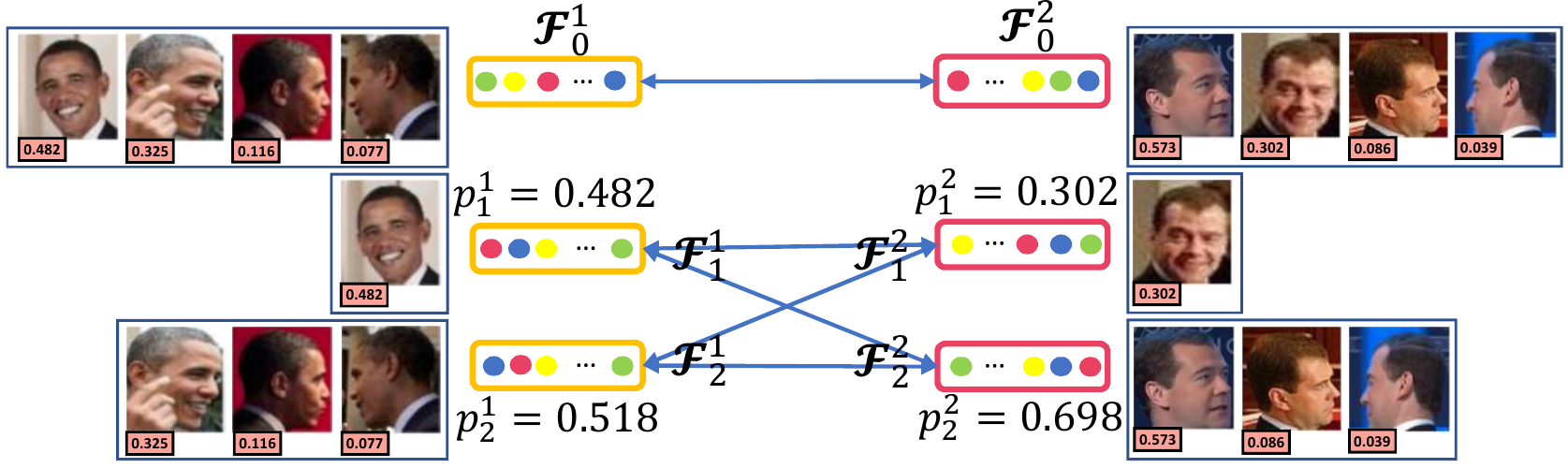}
\caption{Illustration of the parameter-free pose-guided representation scheme. The $p^1_1$, $p^1_2$ are the sum of weights of frontal and profile face images respectively.}
\label{fig:mm6}
\end{figure}

\begin{figure*}[t!]
\centering
\includegraphics[height=7cm]{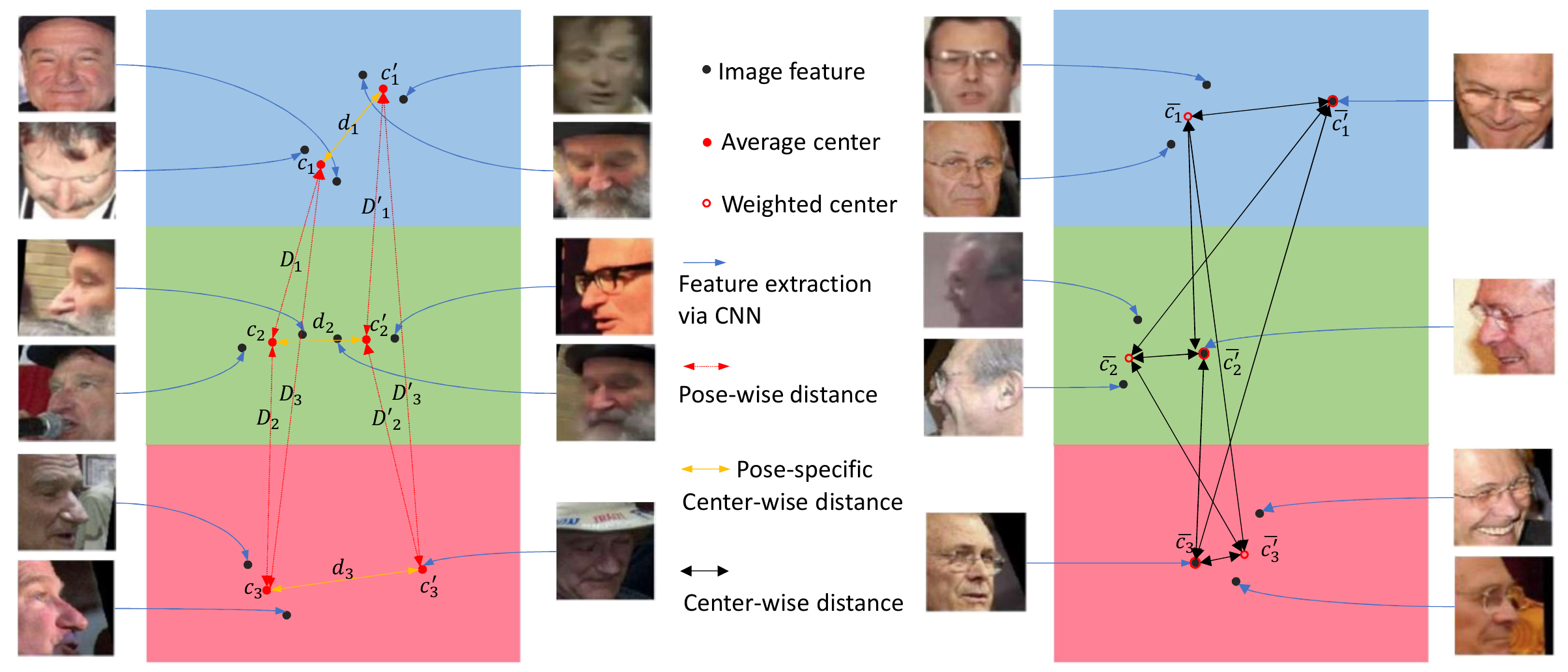}
\caption{ Illustration of the metric learning based pose-guided representation scheme. Left: training stage. Right: testing stage. We use blue, green and red to represent the frontal, left and right face's feature space respectively. The images map the black points via CNN (blue arrows). In training stage (a), the average centers (red points) of each pose group are calculated by average with equal weights, while the weighted average centers (red circle) in testing stage are calculated using the weights assigned by DAC.}
\label{fig:mm7}
\end{figure*}

The other method, temporal convolution, as shown in Fig. 5(b), is introduced as a more efficient alternative for RNN. We apply $64$ convolution kernels of the size $\left\{3,128,1\right\}$ as the first layer and followed by a kernel of $\left\{3,64,1\right\}$ with padding and stripe size is set to 1. The dimension of the output vector is equal to the number of frames in a video. We use softmax to get the normalized attention score. We note that the receptive field of temporal convolution is relatively small compared to the DAC and RNN. However, the redundancy in video is mostly confined to neighboring frames. With this divide-and-conquer idea, we are able to achieve high performance and fewer computations.

\section{Pose-guided inter-set dependency model}

To model the inter-set dependency, we propose a pose-guided stochastic routing scheme. Such an idea originated in \cite{li2006bagging,yin2017multi}, which constructs several face detectors for each view. 

\subsection{Parameter-free PGR}

{We first propose a parameter-free PGR (PF-PGR), which does not use to-be trained parameters.} Given a set of face images, we extract its general feature aggregate $\mathcal{F}_0$, as well as the aggregate of the frontal face features $\mathcal{F}_1$ and profile face feature $\mathcal{F}_2$. $\mathcal{F}_1$ and $\mathcal{F}_2$ are the weighted average of the features from the near-frontal face images ($\leq 30^{\circ}$) and profile face images ($>30^{\circ}$) respectively. We use Pose-invariant face alignment (PIFA) \cite{jourabloo2017pose} to estimate the yaw angle, and train two independent CNNs using the frontal and the profile samples. The sum of weights of the frontal and profile features $p_1$ and $p_2$ reflect the quality of each pose group. We simply mirror the left profile face to the right one. With PGR, the distance $d$ between two sets of samples is computed as:\begin{align}
d= S(\mathcal{F}^1_0,\mathcal{F}^2_0)+ \sum_{i=1}^{2} {\sum_{j=1}^2 {S(\mathcal{F}^1_i,\mathcal{F}^2_j)p^1_ip^2_j}} 
\end{align}where $S$ is the L2 distance between two feature vectors. The distance evaluation complexity is decreased to $\mathcal{O}(5n)$. This achieves promising verification performance requiring fewer comparisons than conventional image-level similarity measurements. It can also be applied to the other variations.

\subsection{Metric learning based PGR}

{Although the PF-PGR is easy for training, pose detection is required in the testing stage, which adds to the processing time. To address this problem, we design a novel metric learning framework (ML-PGR) to align the images in a set with different poses.}

In the training stage, we fine-tune the CNN with the metric learning objective in the training data of IJB series datasets. We separate the samples into frontal, left and right face groups, and calculate their average centers $c_1, c_2$ and $c_3$ respectively. In contrast to the center loss \cite{wen2016discriminative} and (N+M)-tuplet clusters loss \cite{liu2017adaptive,liu2018adaptive,liu2018normalized}, we do not require the samples to be close to their center, as that may lead to loss of their diversity. Instead, we require the different pose groups to be distributed away from each other, and the center of probe-gallery pairs to be near to each other by minimizing the cross-entropy loss of each set and the following loss function with equal weights: \begin{align}
&\sum_{i=1}^{3}max(\beta-D_i,0)^2+\sum_{i=1}^{3}max(\beta-D^{\prime}_i,0)^2\\\nonumber&+\sum_{i=1}^{3}\left\{ (1-l)d_i^2+l\cdot max(\phi-d_i,0)^2\right\}
\end{align} where $D_i$, $D^{\prime}_i$, $d_i$ are the distances of centroids as shown in Fig. 7(a), $\beta$ and $\phi$ are two hyperparameters defining the thresholds of different pose group and probe-gallery pair respectively. $l=0$ if the probe and gallery sets are of the same person, and $l=1$ for different identities. If some pose groups are missing in the probe and gallery sets, the corresponding $D_i$, $D^{\prime}_i$ and $d_i$ are set to 0. The CNN is expected to distribute images in feature space according to their identity and pose. The cross entropy loss can also push different identities to be distributed far away from each other.

We note that the pose groups of gallery samples and their centroids $\bar{c_i}$ can be pre-computed in real-world applications. In the testing stage, we calculate $\bar{c_i}$ in the gallery set using the DAC (or the temporal model for videos). The pose group of each probe image is defined by searching the nearest $\bar{c_i}$, which does not use pose detection. Then, the $\bar{c^{\prime}_i}$ is calculated by the DAC or temporal attention model. The similarity of the probe-gallery pair is defined as the minimum center-wise distance.

\renewcommand{\thefootnote}{\arabic{footnote}}
\section{Numerical Experiments}

\begin{table*}[t!]
\caption{Performance evaluation on the IJB-A dataset. For verification, the true accept rates (TAR) vs. false positive rates (FAR) are reported. For identification, the true positive identification rate (TPIR) vs. false positive identification rate (FPIR) and the Rank-1 accuracy are presented. $~^{\ast}$sd: standard derivation.}
\label{tab:ee1}
\scriptsize 
\begin{center}
\begin{tabular}{|c|c|c|c|c|c|}
    \hline
    \multirow{2}*{Method}& \multicolumn{2}{c|}{1:1 Verification TAR$\pm$sd$~^{\ast}$}& \multicolumn{3}{c|}{1:$N$ Identification TPIR$\pm$sd} \\ \cline{2-6}
    & FAR=0.01 & FAR=0.1 & FPIR=0.01 & FPIR=0.1 & Rank-1 \\ \hline \hline
    B-CNN\cite{chowdhury2016one} & - & - & 0.143$\pm$0.027 & 0.341$\pm$0.032 & 0.588$\pm$0.02\\ \hline
    LSFS\cite{wang2015face} & 0.733$\pm$0.034 & 0.895$\pm$0.013 & 0.383$\pm$0.063 & 0.613$\pm$0.032 & 0.820$\pm$0.024\\ \hline
    DCNN\cite{chen2015end}& 0.787$\pm$0.043 & 0.947$\pm$0.011 & - & - & 0.852$\pm$0.018\\ \hline
   Pose-model\cite{masi2016pose}& 0.826$\pm$0.018 & - & - & - & 0.840$\pm$0.012\\ \hline
   
   Masi~$et~al.$\cite{masi2016we} & 0.886 & - & - & - & 0.906\\ \hline
   Adaptation\cite{crosswhite2017template} & 0.939$\pm$0.013 & 0.979$\pm$0.004 & 0.774$\pm$0.049 & 0.882$\pm$0.016 & 0.928$\pm$0.010 \\ \hline
   QAN\cite{liu2017quality} & 0.942$\pm$0.015 & 0.980$\pm$0.006 & - & - & -\\ \hline
   NAN\cite{yang2017neural} & 0.941$\pm$0.008 & 0.978$\pm$0.003 & 0.817$\pm$0.041 & 0.917$\pm$0.009 & 0.958$\pm$0.005\\ \hline \hline
    DAC(on) & 0.951$\pm$0.014 & 0.980$\pm$0.016 & 0.852$\pm$0.048 & 0.931$\pm$0.012 & 0.970$\pm$0.011\\ \hline
    DAC(off) & 0.953$\pm$0.009 & {0.981}$\pm${0.013}& 0.853$\pm$0.033 & 0.933$\pm$0.006& 0.972$\pm$0.012\\ \hline
    
   { DAC(off)w softmax termination} & {0.938$\pm$0.012 } & { 0.962$\pm$0.009} & {0.806$\pm$0.01} & {0.891$\pm$0.013} & {0.943$\pm$0.014}  \\\hline

    DAC(off)\&PF-PGR &{{0.954}}$\pm${0.01}& {0.981}$\pm${0.008}& {0.855}$\pm${0.042}&{0.934}$\pm${0.009}&{0.973}$\pm${0.011}\\ \hline

    DAC(off)/RNN\&PF-PGR & 0.952$\pm$0.015 & 0.981$\pm$0.012 & 0.854$\pm$0.037 & 0.934$\pm$0.015 & 0.973$\pm$0.015\\\hline
    
    DAC(off)/TempConv\&PF-PGR & {0.960}$\pm$0.012& \textbf{0.986$\pm$0.01} & 0.862$\pm$0.04 & \textbf{0.940$\pm$0.014} & \textbf{0.976$\pm$0.012}\\\hline
    
    DAC(off)/TempConv\&ML-PGR & \textbf{0.963$\pm$0.014} & \textbf{0.986$\pm$0.009} & \textbf{0.865$\pm$0.038} & \textbf{0.943$\pm$0.018} & \textbf{0.976$\pm$0.01}\\\hline
\end{tabular}
\end{center}
\end{table*}

We evaluated the proposed method on three Set/video-based FR datasets: the IJB-A \cite{klare2015pushing}, YTF\cite{wolf2011face}, and Celebrity-1000\cite{liu2014toward}. To utilize the millions of available still images, we train our CNN embedding module separately. As in \cite{yang2017neural}, 3M face images from 50K identities are detected with the Joint cascade face detection and alignment (JDA) \cite{chen2014joint} and aligned using the local binary features (LBF) \cite{ren2014face} method for our GoogleNet training and applied to IJB-A, YTF and celebrity-1000 dataset. The ResNet is pre-trained on VGGFace2 and applied to IJB-B and IJB-C datasets \cite{xie2018comparator}. This part is fixed when we train the DAC module on each set/video face dataset. Benefiting from the highly-compact 128-d feature representation and the simple neural network of the DAC, the training time of our off-policy DAC on IJB-A dataset with a single Xeon E5 v4 CPU is about 3 hours and the average testing time for each set-pair for verification is 62ms. We use Titan Xp for CNN processing. We empirically set $\lambda$ as 0.1, 0.1, 0.1, 0.12 and 0.12 in IJB-A, IJB-B, IJB-C, YTF and Celebrity-1000 respectively. $\beta$ and $\phi$ are set to 1 and 5.

{As our baseline methods, CNN$+$Mean L2 measures the average L2 distances of all image pairs of two sets, which is a decision-level fusion. While the CNN$+$AvePool uses average pooling along each feature dimension for feature -level aggregation. The previous work NAN \cite{yang2017neural} uses the same CNN structure as our framework, but adopts a neural network module for independent quality assessment of each image. Therefore, NAN can be also regarded as a baseline. We refer to the standard A2C as DAC(on), and DAC(off) for the actor-critic with trust region-based experience replay scheme.}

\subsection{Results on IJB-A dataset}

IJB-A \cite{klare2015pushing} is a face $verification$ and $identification$ dataset, containing images captured from unconstrained environments with wide variation in pose and imaging conditions. There are 500 identities with a total of 25,813 images (5,397 still images and 20,412 video frames sampled from 2,042 videos). A set of images for a particular identity is called a template. Each template can be a mixture of still images and sampled video frames. The number of images (or frames) in a template ranges from 1 to 190. It provides a ground truth bounding box for each face with 3 landmarks. There are 10 training and testing splits. Each split contains 333 training and 167 testing identities.

\begin{figure}[t!]

\begin{tabular}{cc}
\includegraphics[height=3.5cm]{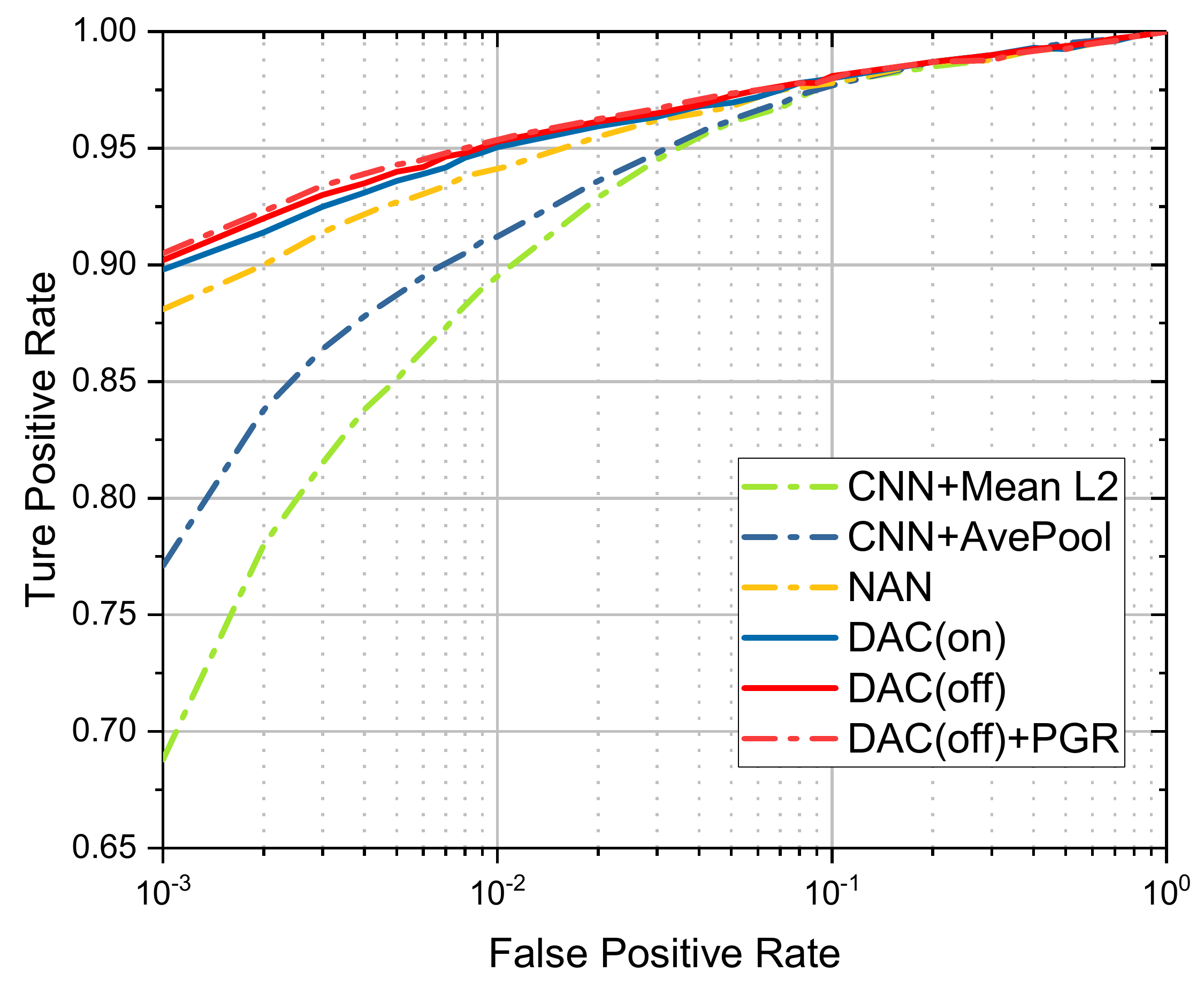}&\includegraphics[height=3.5cm]{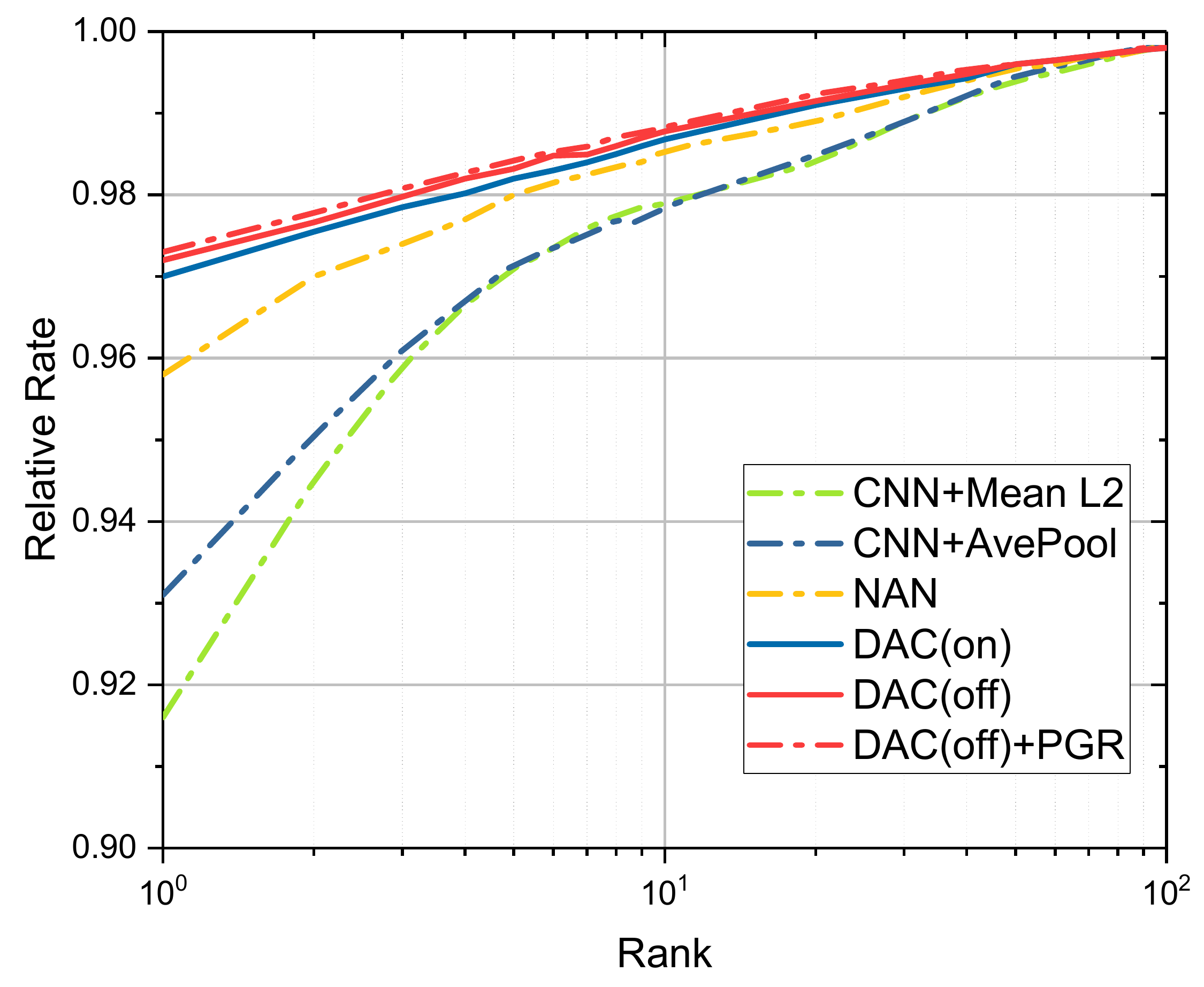}\\
(a)&~~~~~~(b)
\end{tabular}
\caption{Average ROC (Left) and CMC (Right) curves of the proposed method and its baselines on the IJB-A dataset over 10 splits.}
\label{fig:ee8}
\end{figure}

We compare the proposed framework with the existing methods on both face verification and identification tasks following the standard evaluation protocol on IJB-A dataset. For the 1:1 compare task, the receiver operating characteristics (ROC) curves as shown in Fig. 8 (a). We also show the true accept rate (TAR) $vs.$ false positive rates (FAR) in Table I. For the 1:$N$ search task, the Cumulative Match Characteristics (CMC) curve is shown in Fig. 8 (b). A rank-$k$ identification rate is defined as the percentage of probe searches whose gallery match is returned within the top-$k$ matches. The true positive identification rate (TPIR) $vs.$ false positive identification rate (FPIR) as well as the rank-1 accuracy are also reported in Table I.

These results show that both the verification and the identification performance are largely improved compared to baseline methods. The RL networks appear to be robust to low-quality and redundant images. The DAC(on) outperforms the previous approaches at most of the operating points, showing that our representation is more discriminative than the weighted features in \cite{yang2017neural,liu2017quality} without considering the inner-set dependency. {Experience replay can further help the stabilization of our training and results in state-of-the-art performance. Combining the off-policy DAC and pose-guide representation scheme also contributes to the final results. The additional RNN-based temporal attention performs similarly to DAC, while the temporal convolution improves the accuracy significantly. Besides, by dividing the video parts for temporal attention can also reduce the complexity.}

We note that fine-tuning the CNN on IJB-A dataset does not improve the performance, since the image-based datasets are much larger. Compared with \cite{yang2017neural,liu2017quality}, our DAC(off)/TempConv and ML-PGR achieves more than 2\% improvements $w.r.t.$ TAR in FAR=0.01, and 4.8\% $w.r.t.$ TPIP in FPIP=0.01.

{The threshold using the maximum value of softmax predication for termination is validated as 0.6 in IJB-A dataset. Using the new termination condition for the training of IJB-A dataset (which does not have closed-set testing protocol), we note a significant performance drop. This may be because we waste a lot of training samples in the end of image set.}

{In Table. \ref{tab:ee2}, we give the complete comparison of run-time of our proposed methods, and give detailed analysis. We note that the time of CNN to extract 200-dim features using a single Titan Xp GPU is usually about 110mins which processed with GPU. The extracted 200-dim features of each image are storage and our training and testing of RL module is based on that. The feature vector is much small than original images, and our DAC only use two FC layers (100,100) then cascade two sub-branches with two FC-layers (64,16) as shown in Fig. \ref{fig:feed-forward}.Therefore, the memory cost is small, and several images can be computed in parallel.}

{The off-policy used in DAC(off) can efficiently speed up the training, and the trained DAC(off) has similar inference time as DAC(on). These two methods also have similar memory cost in both training and testing. Combining the processing time of CNN extractor, the total training of DAC(off) is 138mins.}

{When combining TempCov or RNN with DAC following the divide and conquer strategy, we use CPU for reinforcement learning module to process order-less images, and run TempCov or RNN in GPU to process video frames. Therefore, the total cost is the slower one. TempConv is more efficient for sequential data, and the DAC(off)/TempConv achieves faster training. The testing time depends on the RL module in CPU. Since DAC only needs to process part of images in a set, its average testing time is 48ms, which is 22\% smaller than DAC(off) only. In contrast, the RNN is usually slower than DAC for both training and testing, the training time depending on the run time of RNN in GPU. Although the RNN takes much more memory than DAC, it uses the GPU memory, which does not affect the DAC using CPU.}

{PF-PGR does not need different training as original DAC, but it requires additional 45ms at testing stage for pose detection and the calculation of Eq. (11). The ML-PGR requires additional training of the feature extractor CNN, but the training time for reinforcement learning part is similar to the DAC(off). In addition, ML-PGR requires retraining of the feature extractor CNN with metric learning objective Eq. (12), which takes about 2hr using a single Titan Xp GPU. The central calculation in ML-PGR will also make its inference time slightly longer than DAC(off), but still much faster than PF-PGR. The memory costs of PF-PGR and ML-PGR are about 1.5$\times$ and 1.1$\times$ of DAC(off) only, since we need to storage two times of features in PF-PGR and the centroids in ML-PGR.}

\begin{table}[t!]
\caption{{Comparison of the average training of the reinforcement learning module and verification time in IJB-A dataset with a single Xeon E5 v4 CPU. *Need retraining the feature extractor CNN.}} 
\label{tab:ee2}
\scriptsize 
\begin{center}
\begin{tabular}{|c|c|c|}
    \hline
    Method&Training Time&Testing Time\\\hline\hline
    DAC(on)&3hr26min&62ms\\\hline
    DAC(off)&2hr18min&62ms\\\hline
    DAC(off)/RNN&4hr23min&136ms\\\hline
    DAC(off)/TempCov&1hr49min&48ms\\\hline
    DAC(off)$\&$PF-PGR&2hr18min&107ms\\\hline
    DAC(off)$\&$ML-PGR&2hr28min*&71ms\\\hline
\end{tabular}
\end{center}
\end{table}

{Our metric learning (ML) based pose guided representation (PGF) is used to remove the pose detection step in testing stage in parameter-free (PF) PGR. Therefore, we can achieve faster inference in testing stage. Note that real-world application usually requires fast testing rather than fast training. We do not manage to get significantly better performance than PF-PGR by using ML-PGR, but the inference time is reduced from 107ms to 71ms.}

{The performance of PGF is closely related to the dataset. The proposed PGR is better suited to handle the probe or the gallery set bias for profile or frontal face. This is a realistic setting, since we might only have a single image or a few images of the same pose of a suspect as probe set.}

{We note that our proposed PGR is helpful in some extreme cases in IJB-A where the probe or gallery set has almost only profile or frontal face, and achieves consistent improvements. For example, in Table I, the TPIR@FPIR=0.01 is improved from 0.853 to 0.855, and the rank-1 accuracy is improved from 0.972 to 0.973, which has been reported in our preliminary ECCV paper. Note that this improvement requires little additional cost. PF-PGR can be added on DAC without any change of the training, and ML-PGR does not need pose detection in testing stage to achieve these improvements.}

{In here we designed a new experimental setting of IJB-A that uses only the profile or frontal face images in the probe set. When the probe set has only profile face images, it is necessary to emphasize the profile face in gallery set, and the frontal face in gallery set is relatively redundant and can be a nuisance factor to the profile faces.}

\begin{table}[t]
\caption{{Open-set identification performance evaluation on the IJB-A dataset with all images/only frontal images/only profile images in the probe set. We note that the ML-PGR is trained with probe set with all images in IJB-A.}}
\label{tab:ee3}
\scriptsize 
\begin{center}
\begin{tabular}{|c|c|c|c|c|}
    \hline
    \multirow{2}*{Probe Set}&\multirow{2}*{Method}& \multicolumn{3}{c|}{1:$N$ Identification TPIR$\pm$sd} \\ \cline{3-5}
    && FPIR=0.01 & FPIR=0.1 & Rank-1 \\ \hline \hline
    IJB-A& DAC(off)&0.853&0.933&0.972\\\cline{2-5}
    Total& DAC(off)$\&$PF-PGR&0.855&\textbf{0.934}&0.973\\\cline{2-5}
    & DAC(off)$\&$ML-PGR&\textbf{0.856}&\textbf{0.934}&\textbf{0.974}\\   \hline 
    
    Only& DAC(off)&0.804&0.883&0.936\\\cline{2-5}
    Frontal& DAC(off)$\&$PF-PGR&\textbf{0.825}&0.906&\textbf{0.958}\\\cline{2-5}
    & DAC(off)$\&$ML-PGR&\textbf{0.825}&\textbf{0.907}&\textbf{0.958}\\   \hline 
    
    Only& DAC(off)&0.776&0.859&0.896\\\cline{2-5}
    Profile& DAC(off)$\&$PF-PGR&0.810&\textbf{0.891}&\textbf{0.930}\\\cline{2-5}
    & DAC(off)$\&$ML-PGR&\textbf{0.811}&\textbf{0.891}&0.929\\   \hline 
    
\end{tabular}
\end{center}
\end{table}

{From Table III, we can see that when only frontal or profile images are available in the probe set, the identification performance drops significantly. The addition of PGR can improve the TPIR@FPIR=0.01/0.1 and Rank-a accuracy by more than 2\% and 3\% for frontal-only and profile-only cases respectively. The ML-PGR does not need pose detection in testing stage to achieve these improvements.}

\subsection{Results on YouTube Face dataset}

The YouTube Face (YTF) dataset \cite{wolf2011face} is a widely used video face $verification$ dataset, which contains 3,425 videos of 1,595 different subjects. This video dataset contains many challenging factors, including amateur photography, occlusions, problematic lighting, pose and motion blur. The length of face videos in this dataset varies from 48 to 6,070 frames, and the average length of videos is 181.3 frames. In experiments, we follow the standard verification protocol as in \cite{yang2017neural,rao2017attention,rao2018learning}, which tests our method for unconstrained face 1:1 verification with the given 5,000 video pairs. These pairs are divided into 10 equal splits, and each split has around 250 intra-personal pairs and 250 inter-personal pairs.

{Table IV} presents the results of our DAC method and previous methods. It can be seen that the DAC outperforms all the previous state-of-the-art methods following the setting without fine-tuning the feature embedding module on YTF. Since this dataset has frontal face bias \cite{crosswhite2017template} and the face variations in this dataset are relatively small as shown in Fig. 2, we did not use the pose-guided representation scheme. It is obvious that the video sequences are redundant, considering the inner-video relationship does contribute to the improvement over \cite{yang2017neural}. The comparable performance with temporal representation-based methods suggests that the DAC could be a potential substitute for RNN in some specific areas. Actually, the RNN itself is also computationally expensive and sometimes difficult to train \cite{zhang2017towards}. We directly model the dependency at the feature-level, which is faster than the Bi-directional LSTM-based model ($e.g.,$ our RNN baseline and \cite{rao2017attention}), and more effective than the adversarial face generation-based method \cite{rao2018learning}. We show that the temporal convolution also promising in video-recognition tasks.

\begin{table}[t!]
\caption{Comparisons of the average verification accuracy with the recently state-of-the-art results on the YTF dataset.$\dag$ fine-tuned the CNN model with YTF. The best results are bolded, and the second best results are underlined.} 
\label{tab:ee4}
\scriptsize 
\begin{center}
\begin{tabular}{|c|c|c|}
   \hline
   Method&Accuracy$\pm$sd&$\dag$Accuracy$\pm$sd\\ \hline \hline
   FaceNet\cite{schroff2015facenet}& 0.9512$\pm$0.0039&-\\
   Deep FR\cite{parkhi2015deep}& 0.915& 0.973\\
   CenterLoss\cite{wen2016discriminative}&0.949&-\\ 
   TBE-CNN\cite{ding2017trunk}&0.9384$\pm$0.0032&0.9496$\pm$0.0031\\
   TR\cite{rao2017attention}&0.9596$\pm$0.0059&0.9652$\pm$0.0054\\
   NAN\cite{yang2017neural}&0.9572$\pm$0.0064&-\\
   DAN\cite{rao2018learning}&0.9428$\pm$0.0069&-\\ \hline \hline
   DAC(on)&0.9597$\pm$0.0041&-\\
   DAC(off)&\underline{0.9601}$\pm$\underline{0.0048}&-\\ 
   RNN&0.9530$\pm$0.0040&-\\
   TempConv&\textbf{0.9680}$\pm$\textbf{0.0038}&-\\ \hline
\end{tabular}
\end{center}
\end{table}

Results also indicate that DAC achieves a very competitive performance without highly-engineered CNN models. Note that the FaceNet \cite{schroff2015facenet}, NAN \cite{yang2017neural} also use the GoogleNet style structure. The Deep FR, TBE-CNN and TR methods have additional fine-tuning of the CNN-model with YTF dataset, and the residual constitutional networks \cite{rao2017attention} are used in TR. Considering our module-based structure, these advanced CNNs can be easily added on the DAC and improve its performance. We see that the DAC can generalize well in video-based face verification datasets.

\begin{figure}[t!]
\begin{tabular}{cc}
\includegraphics[height=3.5cm]{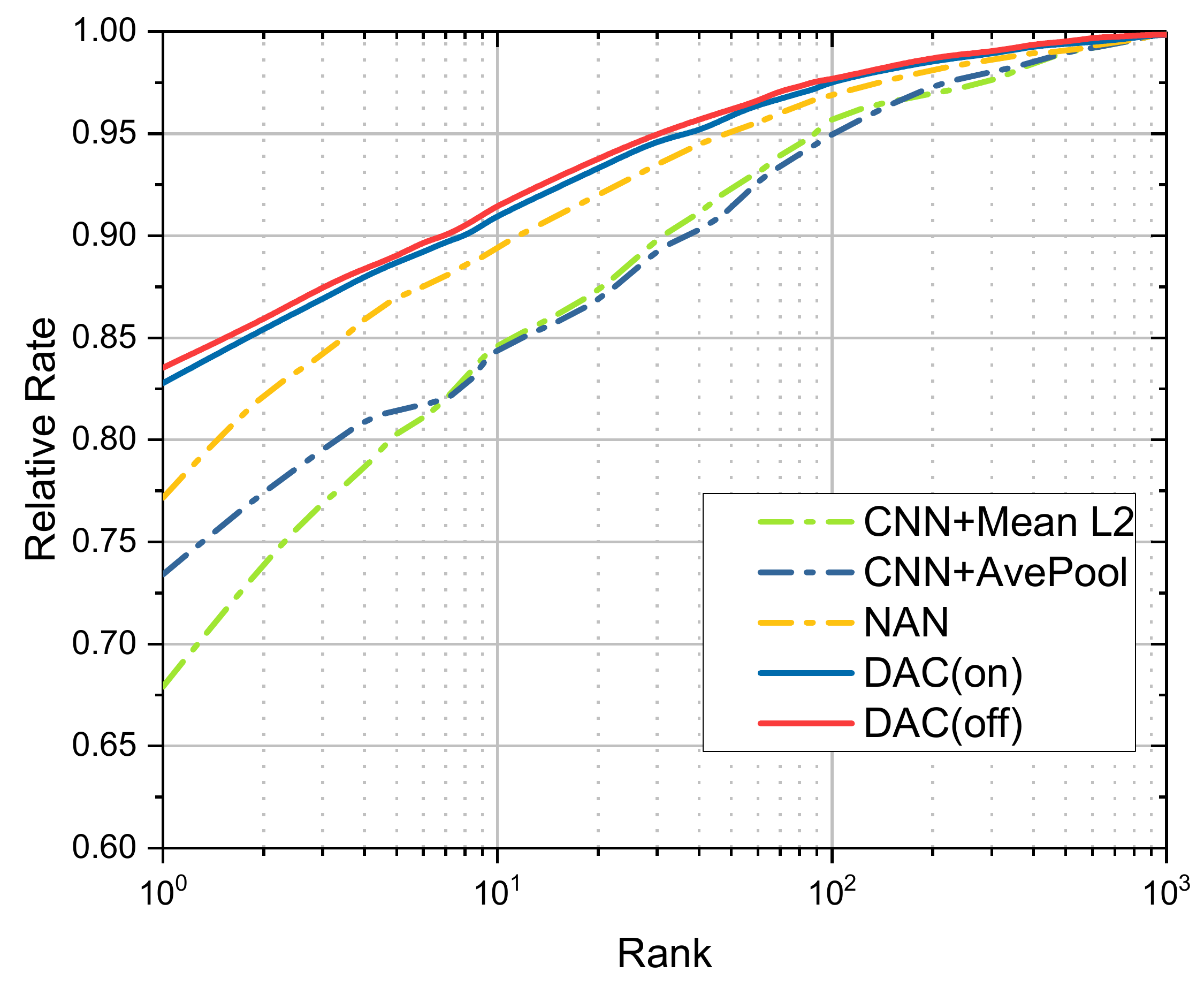}&\includegraphics[height=3.5cm]{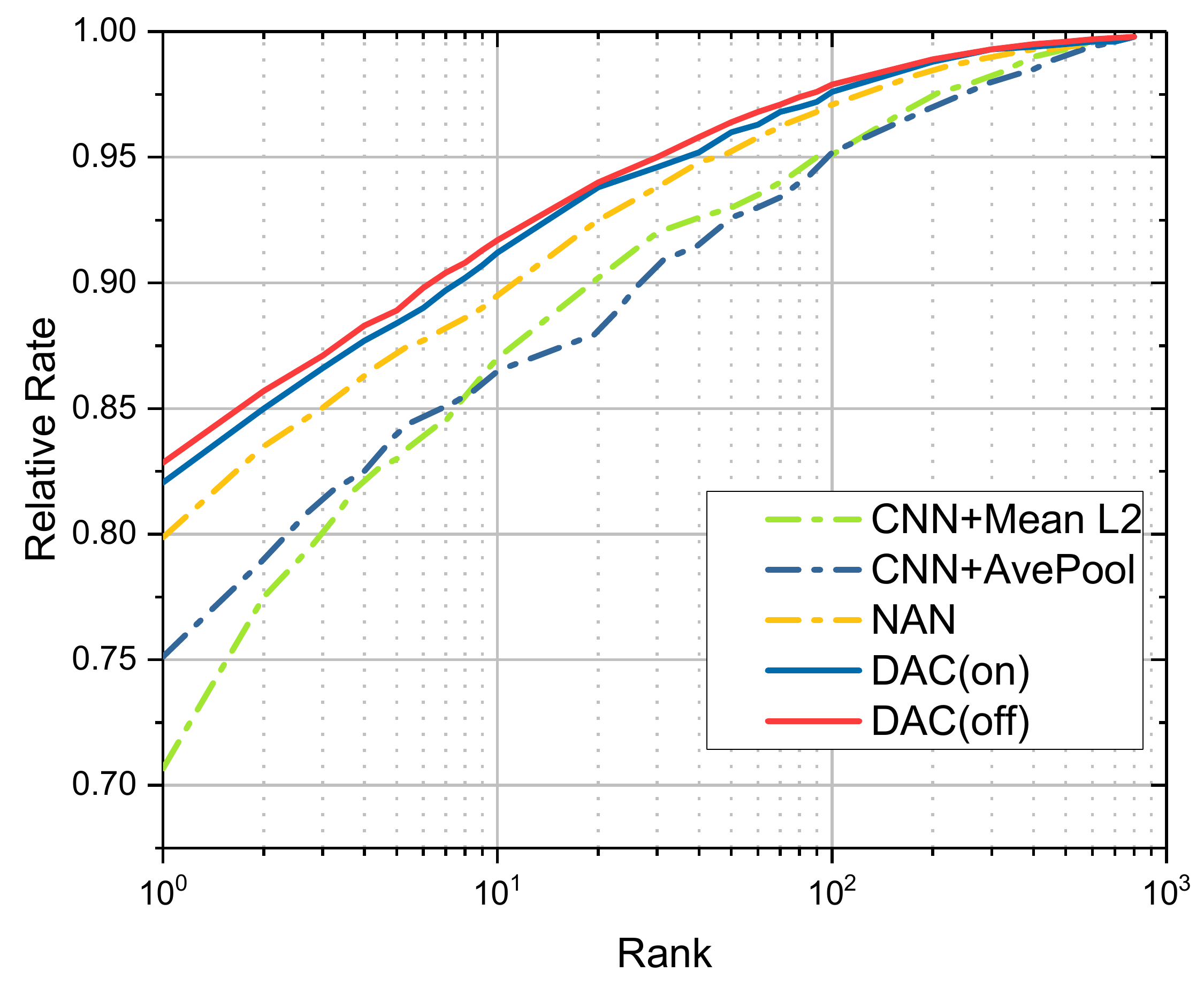}\\
(a)&~~~~~~(b)
\end{tabular}
\caption{
The CMC curves of different methods on Celebrity 1000. (a) Closed-set tests on 1000 subjects, (b) Open-set tests on 800 subjects.}
\label{fig:ee9}
\end{figure}

\begin{table}[t!]
\caption{Rank-1 identification accuracy on the Celebrity-1000 dataset for closed-set tests. The best results are bolded, and the second best results are underlined. *DAC(off) with softmax termination in testing stage.} 
\label{tab:ee5}
\scriptsize 
\begin{center}
\begin{tabular}{|c|c|c|c|c|}
    \hline
    \multirow{2}*{Method} & \multicolumn{4}{c|}{Number of Subjects($closed$)}\\ \cline{2-5}

   & 100 & 200 & 500 & 1000\\  \hline \hline
   MTJSR\cite{liu2014toward}&0.506&0.408&0.3546&0.3004\\ \hline
   Eigen-PEP\cite{li2014eigen}&0.506&0.4502&0.3997&0.3194\\ \hline
   CNN+Mean L2&0.8526&0.7759&0.7457&0.6791\\ \hline
   CNN+AvePool&0.8446&0.7893&0.7768&0.7341\\ \hline
   {CNN+SVM} & {0.8613} & {0.7946} &{0.7792} & {0.7410}\\ \hline
   
   NAN\cite{yang2017neural}&0.9044&0.8333&0.8227&0.7717\\ \hline \hline
   DAC(on)&0.9125&0.8722&0.8475&0.8278\\\hline
   DAC(off)&\underline{0.9137}&\underline{0.8783}&\underline{0.8523}&\underline{0.8353}\\ \hline
   { DAC(off)*} & {0.9135} & {0.8782} &{0.8520} & {0.8349}\\ \hline

   RNN&0.9082&0.8642&0.8265&0.7863\\ \hline
   TempConv&\textbf{0.9160}&\textbf{0.8815}&\textbf{0.8642}&\textbf{0.8661}\\ \hline
   
\end{tabular}
\end{center}
\end{table}

\begin{table}[t!]
\caption{Rank-1 identification accuracy on the Celebrity-1000 dataset for open-set tests. The best results are bolded, and the second best results are underlined.} 
\label{tab:ee6}
\scriptsize 
\begin{center}
\begin{tabular}{|c|c|c|c|c|}
   \hline
    \multirow{2}*{Method}& \multicolumn{4}{c|}{Number of subjects($open$)}\\  \cline{2-5}

      & 100 & 200 & 500 & 800\\  \hline \hline
   MTJSR\cite{liu2014toward}&0.4612&0.3984&0.3751&0.3350\\ \hline
   Eigen-PEP\cite{li2014eigen}&0.5155&0.4615&0.4233&0.2590\\ \hline
   CNN+Mean L2&0.8488&0.7988&0.7676&0.7067\\ \hline
   CNN+AvePool&0.8411&0.7909&0.7840&0.7512\\ \hline
   NAN\cite{yang2017neural}&0.8876&0.8521&0.8274&0.7987\\ \hline \hline
   DAC(on)&0.8986&0.8706&0.8395&0.8205\\ \hline
   DAC(off)&\underline{0.9004}&\underline{0.8715}&\underline{0.8428}&\underline{0.8264}\\ \hline
   RNN&0.8913&0.8642&0.8332&0.8042\\ \hline
   TempConv&\textbf{0.9082}&\textbf{0.8810}&\textbf{0.8613}&\textbf{0.8405}\\ \hline
\end{tabular}
\end{center}
\end{table}

\begin{figure*}[t!]
\centering
\includegraphics[height=4.9cm]{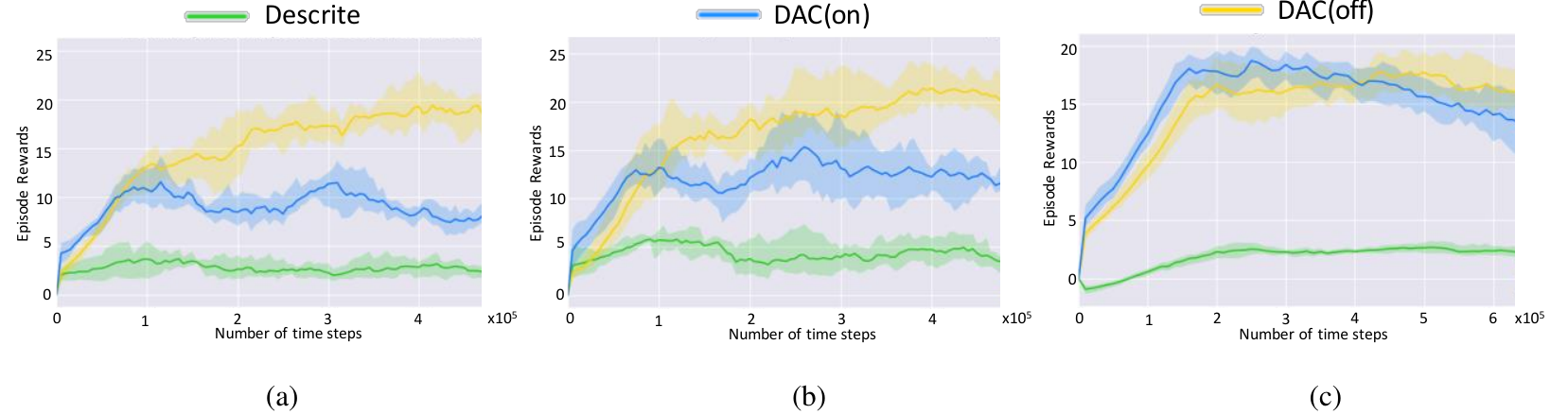}
\caption{Performance comparison on (a) IJB-A, (b) YTF, and (c) Celebrity-1000 datasets. The larger episode rewards, the better training results. The shaded region denotes the standard deviation over 3 random seeds.}
\label{fig:ee10}
\end{figure*}

\begin{table*}
\caption{Performance evaluation on the IJB-B dataset. For verification, the true accept rates (TAR) vs. false positive rates (FAR) are reported.}
\label{tab:ee7}
\scriptsize 
\begin{center}
\begin{tabular}{|c|c|c|c|c|c|}
    \hline
    \multirow{2}*{Method}& \multicolumn{5}{c|}{1:1 Verification TAR}\\ \cline{2-6}
    & FAR=1$E$-5 & FAR=1$E$-4 & FAR=1$E$-3 & FAR=1$E$-2 & FAR=1$E$-1 \\ \hline \hline

    Cao $et al$. \cite{cao2018vggface2} & 0.734 & 0.825 & 0.900 & 0.950 & 0.980\\ \hline
    
    MN \cite{xie2018multicolumn} & 0.771 & 0.862 & 0.927 & 0.968 & 0.989\\ \hline
    
    CN \cite{xie2018comparator} & - & 0.841 & 0.930 & 0.972 & 0.995\\ \hline\hline

    DAC(off)  & 0.811 & 0.869 & 0.938 & 0.972 & 0.995\\ \hline
   
    DAC(off)/TempConv\&PF-PGR & 0.818 & 0.872 & 0.943 & \textbf{0.978} & \textbf{0.996}\\\hline
    
    DAC(off)/TempConv\&ML-PGR & \textbf{0.820} & \textbf{0.874} & \textbf{0.944} & \textbf{0.978} & \textbf{0.996}\\\hline
    
\end{tabular}
\end{center}
\end{table*}

\subsection{Results on Celebrity-1000 dataset}

We also tested our method on the Celebrity-1000 dataset \cite{liu2014toward}, which is designed for the unconstrained video-based face $identification$ problem. 2.4M frames from 159,726 face videos (about 15 frames per sequence) of 1,000 subjects are contained in this dataset. It is released with two standard evaluation protocols: open-set and closed-set. We follow the standard $1:N$ identification setting as in \cite{liu2014toward} and report the result of both protocols.

For the closed-set protocol, we use the softmax outputs from the reward network, and select the subject with the maximum score as the result. Since the baseline methods do not have a multi-class prediction unit, we simply compare the L2 distance as in \cite{yang2017neural}. We present the results in {Table V}, and show the CMC curves in Fig. 9 (a). With the help of end-to-end learning and large volume training data for CNN model, deep learning methods outperform \cite{liu2014toward,li2014eigen} by a large margin. It can be seen that DAC is comparable or better than Bi-directional LSTM, while the temporal convolution usually achieves the best performance in video-based tasks.

{The threshold using the maximum value of softmax predication for termination is validated as 0.5 in Celebrity-1000 dataset. Using the softmax-based terminating condition for the testing stage of closed-set identification in Celebrity-1000 dataset, it achieved a good balance of performance and speed, as shown in Table V. We can see that the rank-1 accuracy of termination according to softmax prediction is comparable to the method that traverse all images. But the average processing times in the testing stage for the case of 100, 200, 500 and 1000 subjects are reduced by 18\%, 15\%, 20\%, 16\% respectively, compared to the method that traverses all the images.}

For the open-set testing, we take multiple image sequences of each gallery subject to extract a highly compact feature representation as in NAN \cite{yang2017neural}. Then the open-set identification is performed by comparing the L2 distance of the aggregated probe and gallery representations. Fig. 9 (b) and {Table VI} show the results of different methods in our experiments. We see that our proposed methods outperform the previous methods again, which clearly shows that DAC is effective and robust.

{We can see from the Tables V and VI and Figs. 8 and 9, that the feature-level AvePool usually outperforms the decision-level Mean L2 by a small margin in IJB-A and Celebrity-1000 datasets. These results are also reported in NAN [58]. We note that the reported CNN+Mean L2 and CNN+AvePool results [58] in closed-set Celebrity-1000 dataset is regarding it as open-set setting.}

{Besides, the decision level fusion with DAC model for image weighting is not effective.  Our DAC is designed for feature-level fusion, since its state is the averaged feature vectors. When we use the weight of an image in probe set times the weight of an image in gallery set as the weight of their L2 distance, and the rank-1 accuracy of DAC(off) using the weighted Mean L2 in IJB-A  is 0.951, which is 2.1\% lower than feature-level fusion DAC(off).}

{The decision-level fusion of SVM classifier is a good choice for the closed-set identification, in which the training and testing have the same subjects. Therefore, the 1 vs rest SVM classifiers trained in the training stage can be used for testing subjects.}

{First, we encode each image using our pretrained feature extractor CNN. Then, the 1 vs rest SVM classifiers are trained using the training set of Celebrity-1000 dataset. At inference stage, we fuse the score of each SMV classifier by majority voting to identify the subject.}

{As we can see in Table V, the performance of 1 vs rest SVM classifiers is better than CNN+Mean L2 and CNN+AvePool, while our DAC(on) outperforms 1 vs rest SVM classifiers by a large margin. The improvement in Rank-1 identification accuracy for 100, 200, 500, 1000 identities setting is 5.12\%, 7.76\%, 6.83\%, 8.68\% respectively. More appealingly, the DAC(off) with RNN or TempConv can achieve even better performance. These results provide further evidence of our contribution in closed-set identification setting.}

\subsection{The reinforcement learning training comparisons}

We show the performance comparisons of DAC(on) and DAC(off) from deep reinforcement learning perspective in three experiments. In addition, we binarized our action space, which means we have drop ($a_t=0$) and keep ($a_t=1$) action choice for each image in a set. This setting is similar to the Q-learning-based cost-efficient approaches. {Fig. 10 shows the episode rewards of each methods. Higher episode rewards are expected for good learning algorithms.} We can see that the continuous action space outperforms the discrete baseline with a large margin in all of the set/video face recognition tasks. {The off-policy actor critic can further achieve higher and more stable convergence results.} Our DAC methods is highly robust to the challenging pose, expressions and imaging conditions in both the set and video face recognition datasets.

{We note that off-policy can also gain improvements in the results of all datasets consistently. Although the gap is relatively small, the improvement of training speed is significant.}

\subsection{Results on IJB-B}

We further tested on IJB-B dataset which is an extension of IJB-A. It collects 21800 still images and 55000 frames in 7011 videos from 1845 subjects. We note that the backbone and pre-training datasets are varied from different papers, we choose the ResNet50 as our CNN and use VGGFace2 for its pre-training in this experiment. The related works that use the same setting are compared in {Table VII}. The improvement of TAR in FAR=1$E$-5 over \cite{xie2018multicolumn} is larger than 4.9\%. A limitation of \cite{xie2018comparator,xie2018multicolumn} is their inefficiency for identification. As shown in {Table VIII}, our method is significantly better than the previous methods $w.r.t.$ Rank-1/5/10 identification accuracy. Compared with Bi-directional LSTM-based image generation method \cite{rao2018learning}, the DAC(off)/TempConv\&ML-PGR achieves 4\% improvement.

\begin{table}
\caption{Performance evaluation on the IJB-B dataset. For identification, the Rank-N accuracy are presented.}
\label{tab:ee8}
\scriptsize 
\begin{center}
\begin{tabular}{|c|c|c|c|}
    \hline
    \multirow{2}*{Method}& \multicolumn{3}{c|}{1:$N$ accuracy$\pm$sd} \\ \cline{2-4}
    & Rank-1 & Rank-5 & Rank-10 \\ \hline \hline
    
    Cao $et~al.$ \cite{cao2018vggface2} & 0.902$\pm$0.036 & 0.946$\pm$0.022 & 0.959$\pm$0.015\\ \hline
    
    DAN \cite{rao2018learning} & 0.899$\pm$0.030 & 0.937$\pm$0.012 & 0.952$\pm$0.012\\ \hline\hline
     
    DAC(off)& 0.928$\pm$0.026 & 0.956$\pm$0.015 & 0.960$\pm$0.018 \\ \hline
   
    DAC(off)/TempConv\&PF-PGR & 0.936$\pm$0.030 & 0.960$\pm$0.017 & \textbf{0.962}$\pm$\textbf{0.015} \\\hline
    
    DAC(off)/TempConv\&ML-PGR & \textbf{0.940}$\pm$\textbf{0.024} & \textbf{0.963}$\pm$\textbf{0.021} & \textbf{0.962}$\pm$\textbf{0.016} \\\hline

\end{tabular}
\end{center}
\end{table}

\subsection{Results on IJB-C}

The IJB-C dataset extends IJB-B, which has 3531 subjects with 31300 still images and 117500 frames from 11779 videos. In its standard verification setting, there are 23124 templates with 15639K imposter matches and 19557 genuine matches. We also follow \cite{xie2018comparator,xie2018multicolumn} to choose ResNet50 as the backbone of CNN and use VGGFace2 for pre-training. The results are shown in {Table IX}.
The DAC(off)/TempConv\&ML-PGR outperforms \cite{xie2018multicolumn} by 6.4\% $w.r.t.$ TAR in FAR=1$E$-5.

\begin{table*}
\caption{Performance evaluation on the IJB-C dataset. For verification, the true accept rates (TAR) vs. false positive rates (FAR) are reported.}
\label{tab:ee9}
\scriptsize 
\begin{center}
\begin{tabular}{|c|c|c|c|c|c|}
    \hline
    \multirow{2}*{Method}& \multicolumn{5}{c|}{1:1 Verification TAR}\\ \cline{2-6}
    & FAR=1$E$-5 & FAR=1$E$-4 & FAR=1$E$-3 & FAR=1$E$-2 & FAR=1$E$-1 \\ \hline \hline

    Cao $et al$. \cite{cao2018vggface2} & 0.647 & 0.784 & 0.878 & 0.938 & 0.975\\ \hline
    
    MN \cite{xie2018multicolumn}  & 0.708 & 0.831 & 0.909 & 0.958 & 0.985\\ \hline
    
    CN \cite{xie2018comparator} & - & 0.880 & 0.944 & 0.981 & 0.998\\ \hline\hline
   
    DAC(off) & 0.754 & 0.882 & 0.958 & 0.985 & 0.998\\ \hline
   
    DAC(off)/TempConv\&PF-PGR & 0.768 & 0.885 & \textbf{0.960} & {0.985} & \textbf{0.999}\\\hline
    
    DAC(off)/TempConv\&ML-PGR & \textbf{0.772} & \textbf{0.886} & \textbf{0.960} & \textbf{0.986} & \textbf{0.999}\\\hline

\end{tabular}
\end{center}
\end{table*}

\section{Conclusion}
We have introduced the actor-critic reinforcement learning (RL) for visual recognition problem. We cast the inner-set dependency modeling to an Markov Decision Process (MDP), and train an agent DAC to achieve attention control for each image in each step. The temporal attention can be easily combined with DAC. {The parameter-free and metric learning-based PGR scheme well balances the computation cost and complementary information utilization.} In the future, we plan to further develop more robust RL algorithm, and following \cite{liu2017adaptive} to learn the thresholds in metric learning adaptively. Although we only explore their ability in set/video-based face recognition tasks, we believe it is a general and practicable methodology that could be easily applied to other problems, such as Re-ID, action recognition and event detection.

%


%

\ifCLASSOPTIONcaptionsoff
  \newpage
\fi



%
\bibliographystyle{ieee}
\bibliography{egbib2}

%







\end{document}